\pdfoutput=1

\documentclass[11pt]{article}

\usepackage[review]{acl}

\usepackage{times}
\usepackage{CJK} 
\usepackage{latexsym}
\usepackage{fontawesome}

\usepackage[T1]{fontenc}

\usepackage[utf8]{inputenc}

\usepackage{microtype}
\usepackage{amsfonts} 
\usepackage{pifont}
\usepackage{cancel}
\usepackage{soul}
\usepackage{xcolor}
\usepackage{algorithm}
\usepackage{algorithmic}
\usepackage[most]{tcolorbox}
\usepackage{booktabs}
\usepackage{xspace}
\usepackage{mathrsfs}
\usepackage{multirow}
\usepackage{amsmath}
\usepackage{colortbl}
\usepackage{subfig}
\usepackage{graphicx}

\definecolor{bleudefrance}{rgb}{0.19, 0.55, 0.91}
\definecolor{yes}{RGB}{239,211,69}
\definecolor{carminered}{rgb}{1.0, 0.0, 0.22}
\definecolor{crimsonglory}{rgb}{0.75, 0.0, 0.2}

\definecolor{second}{HTML}{FFE5D9}
\definecolor{best}{HTML}{FEC89A} 
\definecolor{blueyuhan}{RGB}{0,0,255}

\title{Pastiche Novel Generation: Creating Fan Fiction You Love in \\Your Favorite Author’s Style}

\author{
   Xueran Han\textsuperscript{1}, Yuhan Liu\textsuperscript{2}, Mingzhe Li\textsuperscript{3}, Wei Liu\textsuperscript{4}, Sen Hu\textsuperscript{5}, \\
  \textbf{Rui Yan\textsuperscript{2}, Zhiqiang Xu\textsuperscript{1}, Xiuying Chen\textsuperscript{1}\thanks{\  \
  Corresponding author}} \\
  \textsuperscript{1}MBZUAI, \textsuperscript{2}Renmin University of China, \textsuperscript{3}ByteDance Inc., \textsuperscript{4}Xiaomi, \textsuperscript{5}Peking University\\
}

\begin{document}

\maketitle
\begin{abstract}
Great novels create immersive worlds with rich character arcs, well-structured plots, and nuanced writing styles. 
However, current novel generation methods often rely on brief, simplistic story outlines and generate details using plain, generic language.
To bridge this gap, we introduce the task of \textit{Pastiche Novel Generation}, which requires the generated novels to imitate the distinctive features of the original work, including understanding character profiles, predicting plausible plot developments, and writing concrete details using vivid, expressive language.
To achieve this, we propose WriterAgent, a novel generation system designed to master the core aspects of literary pastiche.
WriterAgent is trained through a curriculum learning paradigm, progressing from low-level stylistic mastery to high-level narrative coherence. 
Its key tasks include language style learning, character modeling, plot planning, and stylish writing, ensuring comprehensive narrative control.
To support this, WriterAgent leverages the WriterLoRA framework, an extension of LoRA with hierarchical and cumulative task-specific modules, each specializing in a different narrative aspect. 
We evaluate WriterAgent on multilingual classics like Harry Potter and Dream of the Red Chamber, demonstrating its superiority over baselines in capturing the target author’s settings, character dynamics, and writing style to produce coherent, faithful narratives.
We hope this work inspires literary creativity in NLP: \faGithub \href{https://anonymous.4open.science/r/writeragent-C0F8/}{WriteAgent}.

\end{abstract}
\section{Introduction}

\begin{figure}[htb]
\centering
\includegraphics[scale=0.35]{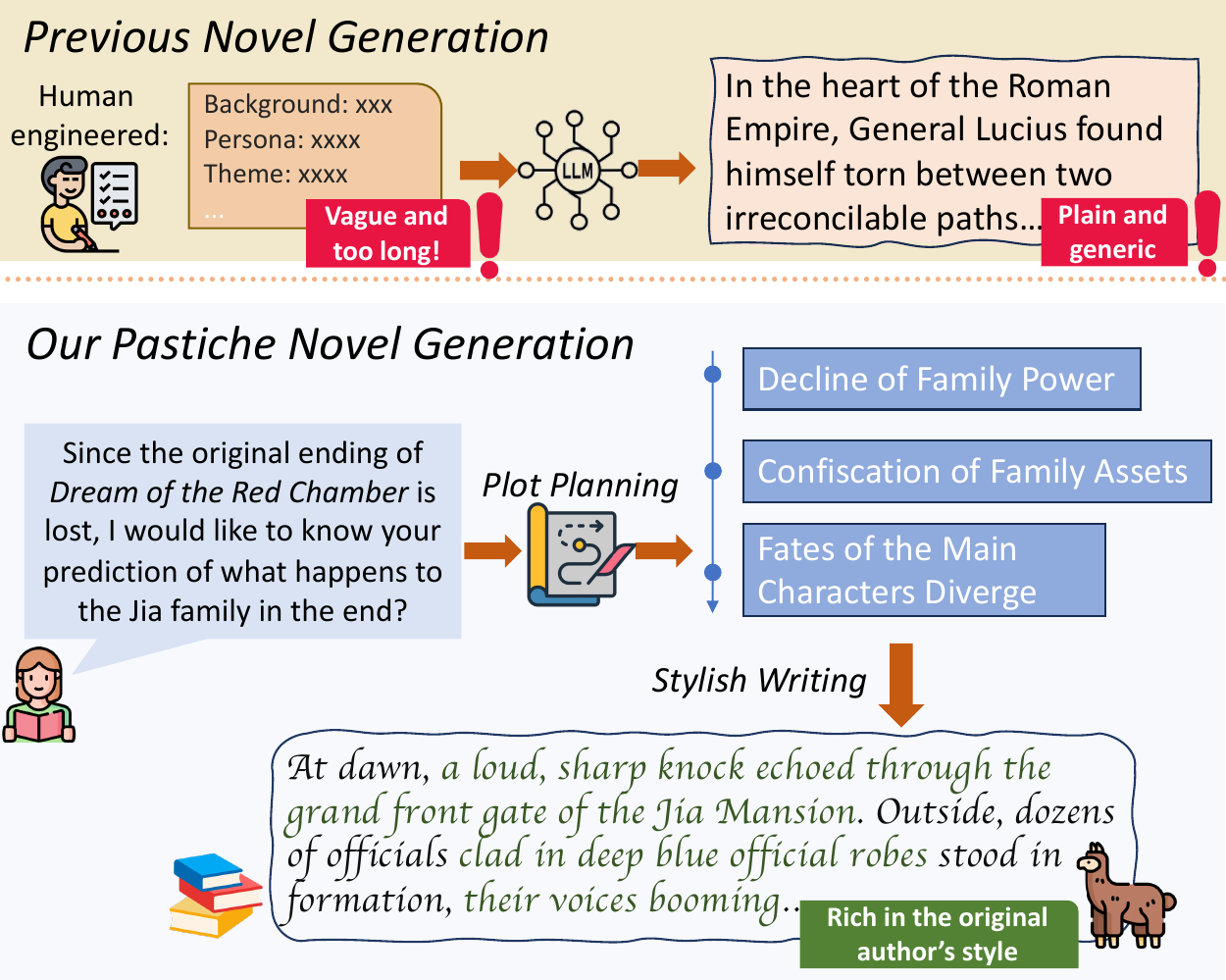}
\caption{
Comparison of traditional story generation and personalized long-form novel generation, showcasing enhanced narrative depth, character development, and stylistic fidelity.
}
\label{fig:compare}
\end{figure}



Novels create rich, immersive worlds with intricate plots and distinct styles, captivating readers through complex storytelling~\cite{bai2024long}. 
A significant amount of research~\cite{ammanabrolu,yao2019plan} has proposed new model architectures to improve story generation. 
With the emergence of LLMs, recent efforts have shifted towards improved prompt-based techniques~\cite{wang2023improving,han-etal-2024-ibsen}. 
For example, \citet{ma2024mops} proposed modular premise synthesis, providing concrete information such as background, persona, and theme to guide the generation process.
While these methods have enhanced novel generation performance~\cite{hu2024novaiterativeplanningsearch}, they fall short in capturing the irreplaceable qualities of real-world literary classics: engaging plots, vivid characters, and distinctive language that immerse readers in complex and authentic storytelling.

Hence, in this work, we propose the \textit{Pastiche Novel Generation} task, which aims to generate novels that faithfully emulate the original author’s style and narrative depth. 
This task presents two key challenges: (1) \textit{plot planning} that aligns with the novel’s established worldview and character dynamics, and (2) \textit{stylish writing} that produces narrative text reflecting the target author’s personalized writing style.
As illustrated in Figure~\ref{fig:compare}, for instance, given the rich context of Dream of the Red Chamber, with its intricate interpersonal conflicts and lavish lifestyles, the model should predict significant plot outcomes, such as the eventual downfall of the Jia family. 
Additionally, the model must accurately reproduce the linguistic and stylistic features of the original text, including evocative phrases like ``rolled up their sleeves'' and ``please issue the decree'', which reflect the author’s unique writing style.

To address these challenges, we propose WriterAgent, a novel generation model designed to emulate a target author's writing style.
The model is trained in sequence on four core tasks to master key narrative and stylistic elements of the author’s work. 
This sequential training follows the natural writing process, from conceptual elements like language style and world-building to high-level plotting and fine details:
1) \textit{Language Style Learning}:
Teaching the LLM to capture the author's distinctive writing style through tasks like next-word prediction, ensuring consistent character voices.
2) \textit{World Building}:
Guiding the model to introduce characters and define their relationships, constructing an interconnected world aligned with the original narrative.
3)~\textit{Plot Planning}:
Enabling the model to generate coherent plotlines that evolve character arcs in line with the story’s structure.
4) \textit{Stylish Writing}:
Enhancing descriptive details of settings, character interactions, and events, ensuring immersive storytelling that reflects the author’s tone and depth.
To train WriterAgent efficiently, we propose WriterLoRA, an extension of the Low-Rank Adaptation (LoRA)~\cite{hu2021lora}.
The original LoRA $A$ and $B$ matrices act as a general expert, preserving the original text’s style as a reference. 
We augment this with specialized $B$ matrices dedicated to specific aspects such as world-building, plot, and detail writing, trained sequentially in a curriculum-based approach from simple to complex tasks.

We evaluated WriterAgent on the English \textit{Harry Potter series} and the Chinese \textit{Dream of the Red Chamber}. 
Given the absence of prior work on the personalized novel generation task, we developed a set of automatic metrics to assess writing style, including language style, expression methods, and sentence complexity, and plot development, covering the story mainline, character behavior, and emotions. 
Both automatic evaluations and human annotations demonstrate that WriterAgent effectively captures the target author’s style and constructs coherent, engaging narratives.

Our main contributions are as follows:
(1) We introduce the Pastiche Novel Generation task, which aims to generate novels that mimic a target author’s style, narrative structure, and character development.
(2) We develop WriterAgent, an LLM with a WriterLoRA structure, trained to learn character profiles, predict future plotlines, and reconstruct full stories, enabling consistent and context-aware storytelling.
(3) We demonstrate that WriterAgent outperforms baseline models in mimicking multilingual classics such as Harry Potter, which highlights new possibilities for literary creativity and personalized storytelling.

\section{Related Work}

\paragraph{Story Generation.}
The task of long-form story generation has received significant attention in recent years due to advancements in LLMs.
Early approaches primarily focused on developing new modules to enhance narrative coherence and consistency.
For example, \citet{ammanabrolu, Fan2019,fan2018hierarchical} leveraged graph structures to organize events more effectively and improve narrative consistency. 
Another example is \citet{peng2018towards}, which introduced an interface for human-computer interaction to generate personalized stories and applied it to RNN-based models for controlling story endings and storylines.
However, these methods often struggled to maintain coherence and consistency over extended~sequences.
More recently, prompt engineering techniques have been adopted to tap into the generative power of LLMs~\cite{giray2023prompt}.
For instance, \citet{han-etal-2024-ibsen} proposed a director-actor agent collaboration framework for controllable and interactive drama script generation, while \citet{huang2023affective} explored dynamic beam sizing and affective reranking to generate engaging narratives.

Despite these advancements, existing methods lack a framework to emulate complex narratives and distinct authorial styles, essential for real-world long-form novel writing.

\paragraph{Parameter-Efficient Fine-Tuning.}

Parameter-efficient fine-tuning (PEFT)~\cite{hetowards} reduces the computational costs of fine-tuning LLMs by introducing additional modules, avoiding direct updates to the large-scale pretrained weights.
Adapters~\cite{houlsby2019parameter} insert extra feature transformations between model blocks, while prefix tuning~\cite{li2021prefix} optimizes parameters through learnable prefixed embeddings without modifying the pretrained weights.
More recently, LoRA~\cite{hu2021lora} introduces low-rank matrix decomposition to efficiently fine-tune models by updating small, trainable parameter matrices. 
Extensions of LoRA have further explored task-specific adapters, enabling specialization in distinct aspects of text generation~\cite{zhou2023path,li2024mixlora, luo2024MoELoRA,,chen2024flexible}.
For example, MoELoRA~\cite{luo2024MoELoRA} adopts a mixture-of-experts (MoE) approach, dynamically selecting different LoRA experts based on the input to improve adaptability. 
Unlike these works, our WriterLoRA combines advantages of a mixture-of-LoRA architecture but is specifically designed to handle the challenges of pastiche novel generation.

\paragraph{Personalized LLMs.}
Personalization is crucial for enabling LLMs to adapt to individual user preferences and requirements. 
Existing approaches to personalization can be broadly categorized into prompt-based methods~\citep{kang2023llmsunderstand,wang2024learningpersonalized} and PEFT-based methods~\citep{cheng2023towards,tan2024democratizing,yu2024neeko}.
Prompt-based methods encode user history and behaviors as contextual examples to guide the model’s response generation~\citep{kang2023llmsunderstand,wang2024learningpersonalized}. In contrast, PEFT-based methods focus on efficiently integrating user-specific preferences into LLMs. For example, \citet{tan2024democratizing} introduced PEFT modules to capture and store user-specific behavior patterns and preferences. Further research has explored improving generalization and efficiency using techniques such as MoE-style gating~\citep{yu2024neeko}, parameter merging strategies~\citep{jang2023personalizedsoup}, and iterative learning~\citep{li2023teachllmspersonalize}.
Building on these foundations, our work extends personalization to stylish novel generation, enabling the model to adapt to complex narrative styles and authorial preferences.

\begin{figure*}[htb]
\centering
\includegraphics[scale=1.1]{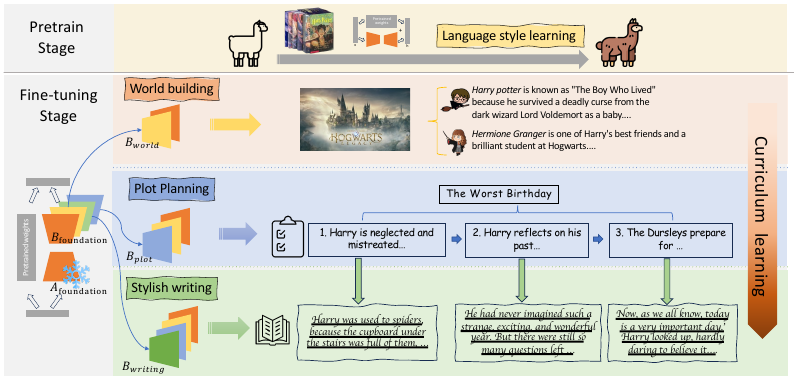}
\caption{
The entire training process can be divided into two parts: the pretraining phase and the fine-tuning phase. During the fine-tuning phase, tasks are divided into three stages of increasing complexity: world-building learning, plot structure learning, and stylish writing learning. These stages are integrated using curriculum learning.
}
\label{fig:framework}
\end{figure*}

\section{Problem Formulation}

We begin by introducing the notations and key concepts for the task of personalized long-form novel generation. 
Formally, the training dataset for a novel is hierarchically structured into character profiles, plots, and words.
Each character profile \( C_i \) consists of the $i$-th character's name and a detailed description, including key traits and relationships with other characters. 
The narrative text is divided into segments of words \( \{x^i_1, x^i_2, \ldots, x^i_n\} \), each corresponding to an individual plot summary \( P_j \), where \( j \) represents the plot index within the chapter. 
These plots \(\{ P_j \}\) capture the key story developments within their respective text segments.

The task involves two core subtasks: (1) \textit{plot prediction}, where the model predicts the next plot \( \hat{P}_t \) based on previous plots \( \{P_1, \ldots, P_{t-1}\} \) and ensures logical consistency by comparing it to the ground truth plot \( P_t \); (2) \textit{stylish writing}, where the model generates a word sequence \( \hat{\mathcal{Y}}_t = \{\hat{x}_1, \hat{x}_2, \ldots, \hat{x}_n\} \) from the predicted plot \( \hat{P}_t \) and ensures coherence and stylistic alignment by comparing it to the ground truth text \(\mathcal{Y}_t = \{x_1, x_2, \ldots, x_n\} \).

\section{Method}

In this section, we first introduce the vanilla LoRA, then present our adapted WriterLoRA built on it, along with the overall WriterAgent framework, as shown in Figure~\ref{fig:framework}.

\subsection{Preliminaries}
LoRA fine-tunes LLMs efficiently by adding trainable low-rank matrices instead of updating all parameters, reducing computational cost.
Concretely, a pre-trained weight matrix  $\mathbf{W}_0 \in \mathbb{R}^{d \times k}$  is updated using a low-rank decomposition  $\mathbf{W}_0 + \Delta \mathbf{W} = \mathbf{W}_0 + \mathbf{B} \mathbf{A}$, where  $\mathbf{B} \in \mathbb{R}^{d \times r}$  and  $\mathbf{A} \in \mathbb{R}^{r \times k}$, with  $r \ll \min(d, k)$.
Here, $\mathbf{B}$ and $\mathbf{A}$ are the trainable low-rank matrices, and $r$ represents the rank of the decomposition.
During training, \( \mathbf{W}_0 \) is frozen and does not receive gradient updates, while \( \mathbf{A} \) and \( \mathbf{B} \) are optimized.
Given an input $\mathbf{x} \in \mathbb{R}^{k}$, the forward pass through the modified weight matrix is:
\begin{align}
    \mathbf{h}' = \mathbf{W}_0 \mathbf{x} + \Delta \mathbf{W} \mathbf{x} = \mathbf{W}_0 \mathbf{x} + \mathbf{B} \mathbf{A} \mathbf{x}.
    \label{update}
\end{align}
Here,  $\mathbf{A}$  serves as an encoder-like transformation, mapping  \( \mathbf{x} \in \mathbb{R}^k\)  into a lower-dimensional representation  $\mathbf{z} \in \mathbb{R}^r$, while  $\mathbf{B}$  acts as a decoder-like transformation, projecting  $\mathbf{z}$  back into output space.

\subsection{Curriculum Learning Tasks}
Our training tasks are inspired by real-world observations of how authors create novels.
Typically, an author begins by determining the work’s style, defining key characters and their traits, outlining interactions, and ultimately developing a complete narrative based on these plots.
Following this natural progression, our model training tasks are designed to emulate this process.
First, the model is pretrained on a next-word prediction task for learning the writing style.
Then, we design three downstream fine-tuning tasks:

\paragraph{World-Building Learning}
In this task, the model maps character names, represented as \( \mathcal{N} = \{N_1, N_2, \ldots, N_n\} \), to their corresponding profiles \( \mathcal{C} = \{C_1, C_2, \ldots, C_n\} \). For each character name \( N_i \), the model generates a detailed profile \( C_i \), including the character's attributes, relationships, and role in the narrative. This process is formulated as:  
\[
    \hat{C}_i = f_{\text{world}}(N_i).
\]  
This task ensures that the model develops a comprehensive understanding of the story world, forming the foundation for subsequent tasks.

\paragraph{Plot Structure Learning}  
Building on the character profiles established in the previous task, this step focuses on predicting narrative progression. The model is trained to generate the next plot \( \hat{P}_t \) based on the most recent \( N_p \) predicted plots:  
\[
    \hat{P}_t = f_{plot}(P_{t-N_p}, ..., P_{t-1}).
\]  
Using the \( N_p \) most recent plots, the model ensures narrative continuity and coherence within the story's timeline.

\paragraph{Stylish Writing}  
Following plot prediction, the Stylish Writing task involves generating the narrative text based on the predicted plot \( \hat{P}_t \).The model produces a sequence of words \( \hat{\mathcal{Y}}_t = \{\hat{x}_1, \hat{x}_2, \ldots, \hat{x}_n\} \) as follows:
\[
    \hat{\mathcal{Y}}_t = f_{\text{writing}}(\hat{P}_t).
\]  
The generated text is then compared with the ground truth \( \mathcal{Y}_t = \{x_1, x_2, \ldots, x_n\} \), ensuring coherence with the plot and stylistic alignment with the author's writing.


\subsection{WriterLoRA}

A straightforward approach to train an LLM involves sequentially training it on the above tasks in a parameter-efficient manner using LoRA. 
However, this method may fail to optimize task-specific performance while maintaining cross-task synergy. 
To address this limitation, we propose WriterLoRA, a structured multi-task learning framework that maximizes efficiency through shared components while ensuring task-specific specialization.  

\paragraph{Shared Foundation}
Initially, the model undergoes next-word prediction training on the entire corpus, using a pair of LoRA matrices, \( \mathbf{A}_{\text{Fdn.}} \) and \( \mathbf{B}_{\text{Fdn.}} \). Here, \( \mathbf{A}_{\text{Fdn.}} \) serves as the shared matrix across all tasks, while \( \mathbf{B}_{\text{Fdn.}} \) collaborates with task-specific \( \mathbf{B} \)-matrices.  
The motivation for sharing the \( \mathbf{A} \)-matrix is to enhance learning efficiency and task synergy.
As a compression matrix, \( \mathbf{A}_{\text{Fdn.}} \) extracts core representations for language understanding and generation, ensuring consistency across tasks while reducing redundancy and improving cross-task transfer learning. 
After pretraining, \( \mathbf{A}_{\text{Fdn.}} \) is fixed to maintain stable representations and prevent catastrophic forgetting.

\paragraph{Task-Specific Adaptations}  
The weight update process originally defined in Equation~\ref{update} is extended to account for task-specific requirements, beginning with language style learning and progressing through subsequent tasks.  
The weight update process defined in Equation~\ref{update} is extended to meet task-specific requirements, starting with language style learning and progressing through world-building, plot structure, and stylish writing tasks.  
During language style learning, the model uses \( \mathbf{B}_{\text{Fdn.}} \) and \( \mathbf{A}_{\text{Fdn.}} \) for next-word prediction, capturing linguistic and stylistic nuances. For world-building, the model maps character names to profiles using \( \mathbf{B}_{\text{world}} \) alongside \( \mathbf{B}_{\text{Fdn.}} \) to define characters and relationships. In plot structure learning, both \( \mathbf{B}_{\text{plot}} \) and \( \mathbf{B}_{\text{world}} \) are used, where \( \mathbf{B}_{\text{plot}} \) controls narrative flow and \( \mathbf{B}_{\text{world}} \) ensures consistency. For stylish writing, \( \mathbf{B}_{\text{writing}} \) integrates with \( \mathbf{B}_{\text{Fdn.}} \), \( \mathbf{B}_{\text{world}} \), and \( \mathbf{B}_{\text{plot}} \) to generate coherent, stylistic text.

The generalized weight update is:  
\[
\resizebox{0.48\textwidth}{!}{
$\mathbf{h}' = \mathbf{W}_0 \mathbf{x} + \left( \mathbf{B}_{\text{Fdn.}} \mathbf{A}_{\text{Fdn.}} + \sum_{t} \alpha_t \mathbf{B}_t \mathbf{A}_{\text{Fdn.}} \right) \mathbf{x}.$
}
\]
The task-specific weights \(\alpha_t\) are computed using:  
\[
\alpha_t = \frac{\exp(w_t)}{\sum_{t'} \exp(w_{t'})},
\]  
where \( w_t = 1 \) for the active task and \( w_{t'} = 0 \) for others. This ensures the active task’s matrix \( \mathbf{B}_t \) dominates, while others provide auxiliary contributions, enabling efficient task adaptation and knowledge sharing.

\begin{table*}[htb]
\small
\centering
\resizebox{\textwidth}{!}{%
\begin{tabular}{l|l|ccc|ccc|ccc}
\toprule[1pt]
\multirow{2}{*}{Dataset} & \multirow{2}{*}{Model} & \multicolumn{3}{c|}{Traditional Metrics} & \multicolumn{6}{c}{Advanced Metrics} \\
                         &                        & ROUGE-1 & ROUGE-2 & ROUGE-L & LS & EX   & SLC  & SM & CBM & EM   \\ 
\hline
\multirow{10}{*}{\begin{tabular}[c]{@{}c@{}}Dream of the\\ Red Chamber\end{tabular}}   
                         & Qwen           & 17.43   & 3.70 & 17.57 & 2.80                     & 2.72 & 2.96 & 2.04 & 2.28 & 2.24  \\
                         & Pretrain(Qwen)    & 20.98 & 5.18 & 18.83 & 2.79                     & 2.68 & \colorbox{second}{3.07} & 2.10 & \colorbox{second}{2.57} & 2.53 \\
                         & LoRA(Qwen)     & \colorbox{second}{31.30} & \colorbox{second}{9.55} & \colorbox{best}{25.58} & \colorbox{second}{3.14}                     & 2.80 & 2.78 & 2.14 & 2.21 & 2.36 \\
                         & MoELoRA(Qwen)        & 23.54  & 6.06 & 16.30 & 3.00                     & \colorbox{second}{2.98}   & 3.02 & \colorbox{second}{2.20} & 2.54 & \colorbox{second}{2.58} \\
                         & \textbf{WriterAgent(Qwen)} & \colorbox{best}{\textbf{34.81}} 
& \colorbox{best}{\textbf{9.73}} 
& \colorbox{second}{\textbf{24.11}}
& \colorbox{best}{\textbf{3.58}} 
& \colorbox{best}{\textbf{3.25}} 
& \colorbox{best}{\textbf{3.27}} 
& \colorbox{best}{\textbf{3.19}} 
& \colorbox{best}{\textbf{3.15}} 
& \colorbox{best}{\textbf{3.12}} \\

\cmidrule{2-10}
                         & ChatGLM         & 15.87 & 2.32 & 13.18 & 2.08                     & 1.96 & 1.96 & 2.19 & 1.87 & 1.79 \\
                          & Pretrain(ChatGLM)         & 15.32 & 2.28 & 13.46 &2.37                     & 2.58 & 2.43 & 2.23 &2.12 & 2.05\\
                         & LoRA(ChatGLM)      & 14.58 & 2.21 & 12.53 & \colorbox{second}{2.54}                     & \colorbox{second}{2.60} & \colorbox{second}{2.87} & 2.26 & \colorbox{second}{2.30} & \colorbox{second}{2.23} \\
                         & MoELoRA(ChatGLM)     & \colorbox{second}{15.89} & \colorbox{second}{2.98} & \colorbox{second}{16.00} & 2.20                     & 2.13 & 2.31 & \colorbox{second}{2.46} & 1.95 & 1.81 \\
                        & \textbf{WriterAgent(ChatGLM) }
& \colorbox{best}{\textbf{16.29}} 
& \colorbox{best}{\textbf{3.21}}
& \colorbox{best}{\textbf{16.01}}
& \colorbox{best}{\textbf{3.03}} 
& \colorbox{best}{\textbf{2.96}} 
& \colorbox{best}{\textbf{2.94}} 
& \colorbox{best}{\textbf{2.59}} 
& \colorbox{best}{\textbf{2.41}} 
& \colorbox{best}{\textbf{2.40}} \\

\hline
\multirow{5}{*}{Harry Potter}   
                         & Llama            & 25.17 & 3.29 & 16.92 & 2.55                        & \colorbox{second}{2.64}    & 3.03    & 3.03    & \colorbox{second}{3.04}    & \colorbox{best}{2.85} \\
                         & Pretrain(Llama)       & \colorbox{second}{26.30} & 4.29 & 18.71 & 2.42                        & 2.32    & 2.95    & 2.20    & 2.31    & 2.26 \\
                         & Lora(Llama)           & 26.18 & 6.35 & 18.97 & 2.56                        & 2.53    & 2.95    & 2.53    & 2.46    & 2.45 \\
                         & MoELoRA(Llama)        & 25.43 & \colorbox{second}{7.93} & \colorbox{second}{20.23} & \colorbox{second}{2.68}                        & 2.58    & \colorbox{second}{3.15}    & \colorbox{second}{3.23}    &2.95    & 2.53 \\
                         & \textbf{WriterAgent(Llama)} & \colorbox{best}{\textbf{29.24}} 
& \colorbox{best}{\textbf{8.79}}
& \colorbox{best}{\textbf{23.91}}
& \colorbox{best}{\textbf{3.05}} 
& \colorbox{best}{\textbf{2.86}} 
& \colorbox{best}{\textbf{3.29}} 
& \colorbox{best}{\textbf{3.54}} 
& \colorbox{best}{\textbf{3.03}} 
& \colorbox{second}{2.75} \\

\bottomrule[1pt]
\end{tabular}}
\caption{Performance comparison of baseline models and our WriterAgent on two classic literary datasets, Harry Potter (English) and Dream of the Red Chamber (Chinese), evaluated for personalized long-form novel generation.
The scores are represented as follows: \colorbox{best}{best} and \colorbox{second}{second}. 
Numbers in \textbf{bold} mean that the improvement to the best baseline is statistically significant (a two-tailed paired t-test with p-value \textless 0.01).}
\label{table:main}
\end{table*}

\section{Experimental Setup}
\label{sec:experiment}

\subsection{Dataset}

We selected two renowned literary works for our dataset: the classic Chinese novel Dream of the Red Chamber and the Harry Potter series, chosen for their rich narratives and literary significance.  
For Dream of the Red Chamber, the first 80 chapters were used for training and the last 40 for testing. Similarly, the first six Harry Potter books were used for training, with the final book for testing.

In addition to the primary text, our dataset incorporates supplementary information to enhance its utility. 
First, we collected detailed, human-written \textit{introductions for the main characters}. 
These character profiles provide valuable context for tasks such as role-playing and characterization. 
Specifically, profiles for Dream of the Red Chamber were sourced from ~\href{https://www.sohu.com/a/773246248_121948389}{Sohu website}, while those for Harry Potter were obtained from~\href{https://www.wikiwand.com/en/List_of_Harry_Potter_characters}{Wikipedia}. 
Secondly, we used GPT-4 to segment the text into sections and generate concise \textit{plot summaries} for each section.  
A section is smaller than a chapter but longer than a paragraph, with the division based on self-contained and relatively complete narrative events.  
These summaries offer structured descriptions of key narrative developments, supporting tasks such as plot-aware content generation.

\subsection{Comparison Methods}

We selected ChatGLM and Qwen as backbone models for evaluating Chinese language performance, and Llama3 for English tasks. 
The key models included in our comparisons are as follows:
(1) \textit{Qwen2-7B-Instruct}~\cite{bai2023qwen}: An instruction-tuned variant of Qwen2-7B, optimized for handling complex queries and interactive tasks.
(2) \textit{ChatGLM2-6B}~\cite{zhang2023chatglm}: A bilingual LLM optimized for both English and Chinese languages.
(3) \textit{Llama3-8B}~\cite{touvron2023llama}: Excels in reasoning, creative writing, and coding tasks.
We also include fine-tuned LLMs tailored for novel generation:   
(4) \textit{LoRA}~\cite{hu2021lora}: An efficient fine-tuning method applying low-rank updates without modifying all model parameters.  
(5) \textit{MoELoRA}~\cite{liu2024moe}: Combines LoRA with MoE architecture, activating subsets of parameters. MoELoRA uses three pairs of $A$ and $B$ matrices, each specialized for world building, plot learning, or stylish writing.

\subsection{Implementation Details}  
We implemented our experiments using PyTorch and conducted them on an NVIDIA A100 GPU. All models were configured with a maximum sequence length and cutoff length of 2048 tokens.
We trained the model for 3 epochs using the AdamW optimizer and BF16 precision for efficiency. The learning rate was set to 1.0e-4 and scheduled using a cosine decay strategy with a warmup ratio of 0.1.
A dropout rate of 0.05 was applied to prevent overfitting. 
The rank \( r \) was set to 8 for both LoRA and each LoRA module within MoELoRA.  
To simulate a larger batch size, we set the batch size to 1 and applied gradient accumulation over 8 steps. More implementation details are in Appendix~\ref{imple}.

\subsection{Evaluation Metrics}  


\paragraph{Novel Generation Evaluation:}
The primary goal of our task, novel generation, is evaluated using both traditional metrics and six advanced aspect-based metrics.
Traditional metrics like ROUGE-N and ROUGE-L~\cite{lin2004rouge} assess content coverage, fluency, and structural alignment. ROUGE-N measures n-gram overlap, while ROUGE-L evaluates the longest common subsequence to capture broader structural coherence.

\begin{table}[tb]
\small
\centering
\resizebox{0.5\textwidth}{!}{%
\begin{tabular}{l|l|ccc}
\toprule[1pt]
\multirow{2}{*}{Dataset} & \multirow{2}{*}{Model} & \multicolumn{3}{c}{Advanced Metrics} \\
                         &                        & SM   & CBM   & EM    \\ 
\hline
\multirow{10}{*}{\begin{tabular}[c]{@{}c@{}}Dream of the\\ Red Chamber\end{tabular}}   
                         & Qwen           & 1.89                     & 2.17 & 2.18  \\
                         & Pretrain(Qwen)    & 1.94                     & 2.13 & 2.26  \\
                         & LoRA(Qwen)     & 1.97                     & 2.14 & 2.29  \\
                         & MoELoRA(Qwen)        & 2.01                     & 2.23   & 2.35  \\
                         & \textbf{WriterAgent(Qwen)} 
& \colorbox{best}{\textbf{2.14}} 
& \colorbox{best}{\textbf{2.54}} 
& \colorbox{best}{\textbf{2.78}} \\

\cmidrule{2-5}
                         & ChatGLM         & 1.78                     & 1.56 & 1.43  \\
                         & Pretrain(ChatGLM)         & 1.86                     & 1.75 & 1.53  \\
                         & LoRA(ChatGLM)      & 1.80                     & 1.79 & 1.60  \\
                         & MoELoRA(ChatGLM)     & 1.88                     & 1.83 & 1.72  \\
                         & \textbf{WriterAgent(ChatGLM)} 
& \colorbox{best}{\textbf{1.86}} 
& \colorbox{best}{\textbf{1.91}} 
& \colorbox{best}{\textbf{1.82}} \\

\hline
\multirow{5}{*}{Harry Potter}   
                         & Llama            & 1.97                        & 1.86    & 1.49    \\
                         & Pretrain(Llama)       & 2.06                        & 1.95    & 1.56    \\
                         & LoRA(Llama)           & 2.03                        & 2.01    & 1.76    \\
                         & MoELoRA(Llama)        & 2.08                        & 2.11    & 1.85    \\
                         & \textbf{WriterAgent(Llama)} 
& \colorbox{best}{\textbf{2.10}} 
& \colorbox{best}{\textbf{2.13}} 
& \colorbox{best}{\textbf{1.87}} \\

\bottomrule[1pt]
\end{tabular}}
\caption{Scores for plot prediction ability, with \colorbox{best}{best} highlighting and \textbf{bold} denoting statistically significant improvement (p-value \textless 0.01).}
\label{table:plot}
\end{table}

Beyond these traditional measures, we introduce six advanced aspect-based metrics to evaluate both \textit{\textbf{writing style}} and \textit{\textbf{plot development}}. 
These aspects are derived from human observations of a randomly selected set of 100 samples.
Writing style evaluation includes:
\textit{Language Style (LS)}, which assesses similarity in vocabulary choices, sentence structures, tone, and voice;
\textit{Expression Methods (EX)}, which evaluates consistency in emotional expression, descriptive techniques, and use of metaphors;
\textit{Sentence Length and Complexity (SLC)}, which compares sentence structures, their complexity, and overall paragraph organization.
Plot development evaluation includes:
\textit{Story Mainline (SM)}, which measures the alignment and coherence of the central plotline;
\textit{Character Behavior and Motivation (CBM)}, which examines whether character actions and motivations are logically consistent with the story’s progression;
\textit{Emotions (EM)}, which evaluates the emotional flow and conflict dynamics within the text.
Given the high correlation of GPT-4o with human judgments, particularly for creative NLG tasks~\cite{wang2023chatgpt}, we adopt both GPT-4o-based scoring and human evaluation. 
GPT-4o rates each generated text from 1 to 5, based on comparison with the ground truth.
For the full content, please refer to Appendix~\ref{prompts}

\begin{figure*}[htb]
\centering
\includegraphics[width=1\textwidth]{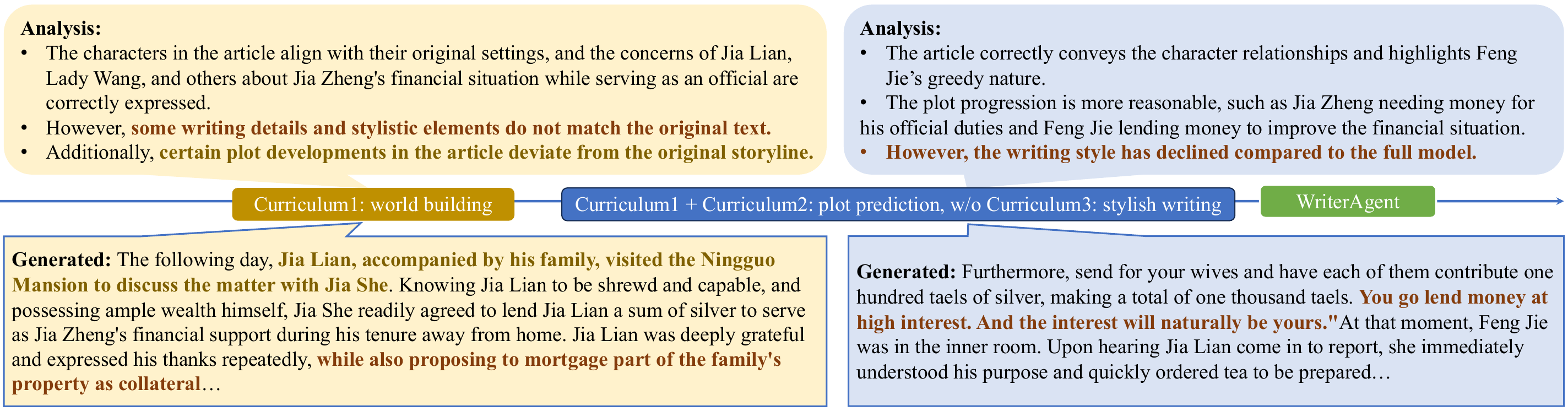}
\caption{
Demonstration of the model's stepwise learning on Dream of the Red Chamber: from curriculum 1 to curriculum 1 \& 2. The text colors indicate the corresponding problems.
Chinese version is in Appendix~\ref{fig:step_chinese}.
}
\label{fig:step}
\end{figure*}



\begin{figure}[htb]  
\centering  
\includegraphics[width=0.45\textwidth]{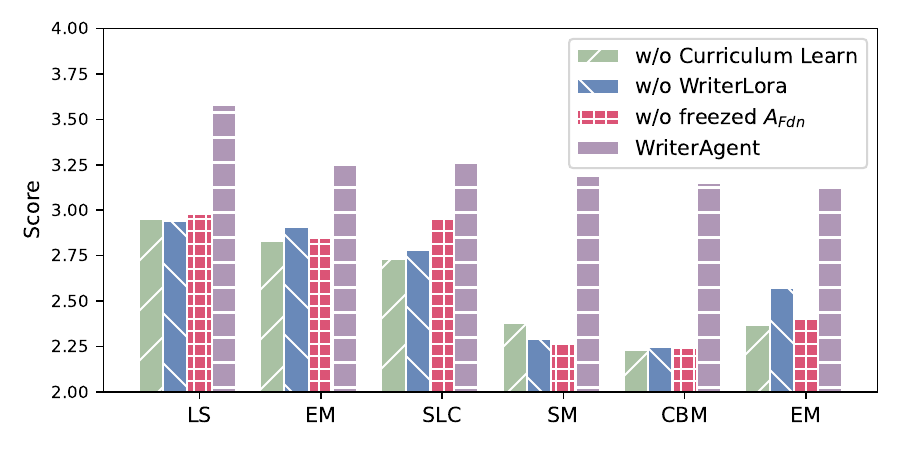}  
\caption{Ablation performance of WriterAgent.}  
\label{fig:ablation}  
\end{figure}

\paragraph{Plot Planning Evaluation:}  
As an intermediate step crucial to creating a well-structured and engaging narrative, plot planning is evaluated using a subset of advanced aspect-based metrics centered on plot coherence. Specifically, \textit{SM}, \textit{CBM}, and \textit{EPM} are applied. The evaluation follows the same GPT-4o-based scoring setup as described earlier.

\section{Experimental Results}
\subsection{Main results}
Table~\ref{table:main} and Table~\ref{table:plot} present the experimental results of Qwen-2, Pretrain, SFT, MoELoRA, and WriterAgent across two datasets.

Firstly, \textit{evaluation results reveal that different metrics capture complementary aspects of writing quality.}
ROUGE favors models like LoRA for content coverage and structural alignment, while aspect-based metrics highlight Pretrain and MoELoRA’s strengths in plot coherence and emotional tone. This underscores the need for diverse evaluation frameworks to fully assess model performance.
Secondly, MoELoRA, the best-performing baseline, benefits from specialized multi-task training, producing coherent plots and well-aligned emotional tones, but \textit{its language style is not well-preserved due to the loss of pre-trained textual style during multi-task learning}. 
Finally, our proposed \textit{WriterAgent consistently outperforms all baselines across datasets and metrics}, achieving significant improvements in plot coherence, character development, emotional depth, and overall narrative quality. 
Additionally, it addresses MoELoRA’s language style limitation by freezing the matrix $A$, thereby preserving the pre-trained writing style and ensuring stylistic consistency in the generated narratives.

\begin{figure*}[htb]
\centering
\includegraphics[scale=0.35]{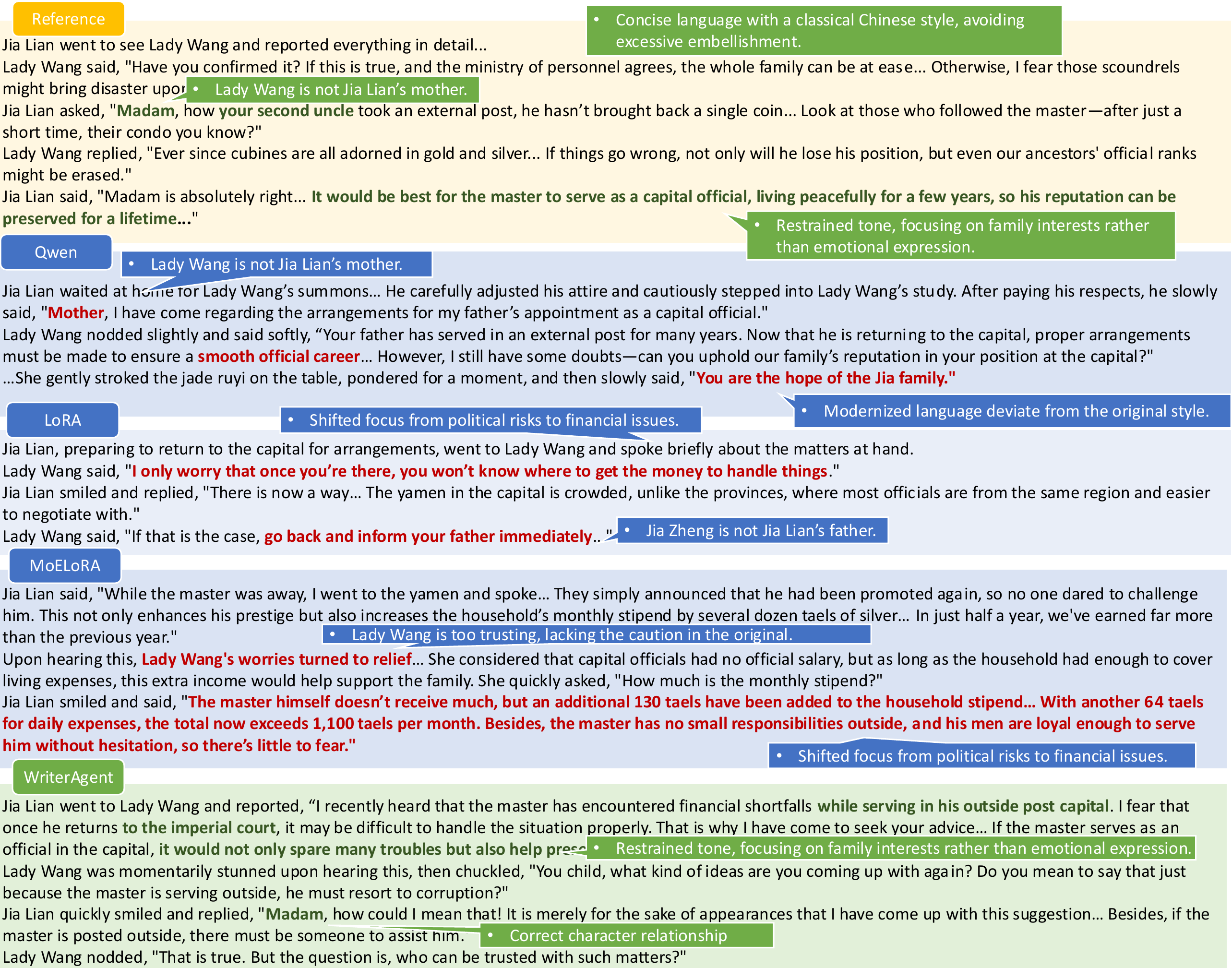}
\caption{
Comparison of reference and generated texts from the baseline and our WriterAgent.
We highlight the weaknesses of the baseline model and the strengths of our approach. 
Some linguistic nuances may be lost in translation; see Figure~\ref{fig:write_red_chinese} in the Appendix (Chinese) for accuracy.
}
\label{fig:write_red}
\end{figure*}

\subsection{Ablation Study}

We conduct an ablation study on Dream of the Red Chamber to evaluate the impact of different learning strategies and LoRA configurations in our model, as shown in Figure~\ref{fig:ablation}. 
Removing the curriculum learning strategy led to a significant performance drop across all metrics, showing that learning all tasks simultaneously without a structured progression reduces the model’s effectiveness.  
Similarly, removing WriterLoRA with only one LoRA weakened task-specific adaptation, reducing the model’s ability to handle diverse constraints. 
Finetuning the foundation matrix \(A_{fdn.}\) instead of keeping it fixed led to performance degradation, showing the importance of a stable foundation for consistent improvements.

\subsection{Analysis of Curriculum Learning}
Our curriculum learning consists of three stages: world building, plot prediction, and stylish writing. Figure~\ref{fig:step} presents a case study showing the model's output after each stage.
After curriculum 1 (world building), the generated text demonstrates characters and traits that closely align with the original work’s character settings. However, the plot structure and textual details remain significantly different from the original. 
After curriculum 2 (plot prediction), the output retains the vivid character traits learned in the first stage. 
For instance, Wang Xifeng is depicted as highly fond of wealth and enthusiastic about managing the household, consistent with her original portrayal. 
However, since the plot construction relies on modern vernacular organization, the model's ability to capture writing details and mimic the original literary style is insufficient.
Finally, after incorporating the final step of curriculum learning, the overall performance improves significantly in terms of word choice, plot structure, and style, as shown in Figure~\ref{fig:write_red}. We present Chinese output in Figure~\ref{fig:step_chinese}.

\subsection{Human Evaluation}  
In addition to automatic evaluation, we conducted a human evaluation with two PhD annotators, both native speakers with strong literary backgrounds. 
They select the best-performed model on the Dream of the Red Chamber dataset.
The results indicate that our model was preferred in 73.1\% of cases, achieving a Kappa score of 0.78, which signifies substantial agreement.
See Figure~\ref{fig:pie} in the Appendix for details.

In Figure~\ref{fig:write_red}, we present a case study of the generated novel text and categorize common errors in baseline models.
First, \textit{\textbf{character relationship}} errors frequently occur, leading to misinterpretations of key familial roles. 
For instance, baselines such as Qwen and LoRA misidentify Lady Wang as Jia Lian’s mother, distorting the intricate family dynamics in Dream of the Red Chamber.
Second, \textit{\textbf{plot inconsistencies}} arise, where baselines shift core narrative elements. 
For example, LoRA and MoELoRA prioritize financial matters over official duties, altering the intended thematic focus and disrupting the novel’s structured storytelling.
Third, \textit{\textbf{stylistic deviations are prevalent}}, as baselines modernize language, losing conciseness and tradition; for example, Qwen adds ``hope of the Jia family'', disrupting the reserved tone.
Additionally, errors in emotional tone make dialogues overly sentimental and expressive, diverging from the novel’s subtle and restrained emotional depth.
In contrast, our model ensures plot consistency, preserves character relationships and power dynamics, and maintains classical language and restrained emotion, staying true to Lady Wang’s focus on political risks and family reputation.
We present further analysis in Appendix~\ref{appendix_red}, with Harry Potter cases in Appendix~\ref{harry}.

\section{Conclusion}

We introduce WriterAgent, an LLM designed for novel generation by learning character profiles, predicting plotlines, and reconstructing full stories. Unlike traditional models reliant on prompt engineering, WriterAgent internalizes novel-related knowledge, reducing external dependencies.
At its core, WriterLoRA employs LoRA modules and curriculum learning to progressively master narrative elements, enhancing coherence and stylistic consistency.
Experiments on two datasets show that WriterAgent outperforms baselines in capturing complex settings and character dynamics. In the future, we aim to extend WriterAgent’s capabilities to broader literary genres and 
refine its adaptability across diverse storytelling styles.


\section*{Limitation}
Despite the effectiveness of WriterAgent in generating pastiche novels, several limitations remain.
First, while determining the plot before drafting is a natural part of the writing process, plots typically exist in the author's mind rather than as explicit, structured data. 
As a result, we need to manually construct plot datasets, which introduces potential biases and may not fully capture the organic, evolving nature of storytelling.
Second, automatic evaluation heavily relies on LLMs' ability to understand an author's style, yet current LLMs do not possess comprehensive knowledge of all literary nuances. 
To mitigate this, we incorporated few-shot prompting in our evaluation, allowing the model to refine its understanding of specific authors. 
However, this approach is still limited and cannot fully replace human judgment, as LLM-based evaluation may overlook deeper stylistic elements and narrative coherence that human readers naturally perceive.
These limitations highlight the challenges in literary pastiche generation and evaluation, underscoring the need for future research on more refined plot modeling and improved evaluation frameworks that better align with human literary perception.

\paragraph{Ethical Considerations}
The development of WriterAgent involves certain ethical considerations, primarily regarding intellectual property, authenticity, and responsible use. Our goal is to learn and emulate an author’s style rather than directly replicate their work, and we prioritize fairness and compliance when working with copyrighted texts. Additionally, generating text that closely resembles an author’s writing style may raise discussions about authenticity. Therefore, we encourage responsible usage and advocate for AI to function as an assistive tool in a collaborative creative environment, fostering literary diversity while respecting the original authors' writing styles and creative spirit. To this end, we can implement measures such as limiting outputs to derivative or transformative works, providing transparency about AI-generated content, and ensuring that the system serves as a tool for creative inspiration rather than a substitute for human authorship.

\bibliography{acl_custom}

\clearpage
\appendix

\section{Implementation Details}
\label{imple}

The training progress was monitored every 10 steps, with checkpoints saved every 500 steps to enable recovery and evaluation.
Loss curves were plotted to track convergence, with the output directory overwritten during updates to maintain consistency. To improve efficiency, data preprocessing utilized 8 parallel workers. During training, we set the batch size to 1 and used gradient accumulation over 8 steps to simulate a larger batch size. The number of samples was limited to 1000 to control training time. The training dataset followed the Alpaca format, and for the multi-task MoE model, a \textit{task\_id} field was added to the data to classify tasks during training.

We used the AdamW~\cite{loshchilov2017decoupled} optimizer for training, which decouples weight decay from gradient updates, improving both training stability and generalization. This makes it particularly effective for large-scale models, such as Transformers, by avoiding overfitting and enhancing optimization efficiency.

The version of the Transformers library was chosen to match the architectures of the models being trained. For the ChatGLM2-6B model, which adopts a GLM (General Language Model) architecture with bidirectional and autoregressive training, version 4.30.2 of Transformers was used. However, for the Llama 3-8B and Qwen2-7B-Instruct models, the recommended version was 4.45.0 to ensure compatibility and optimal performance.

\section{Case Study of Harry Potter }
\label{harry}

\begin{figure*}[htb]
\centering
\includegraphics[scale=0.35]{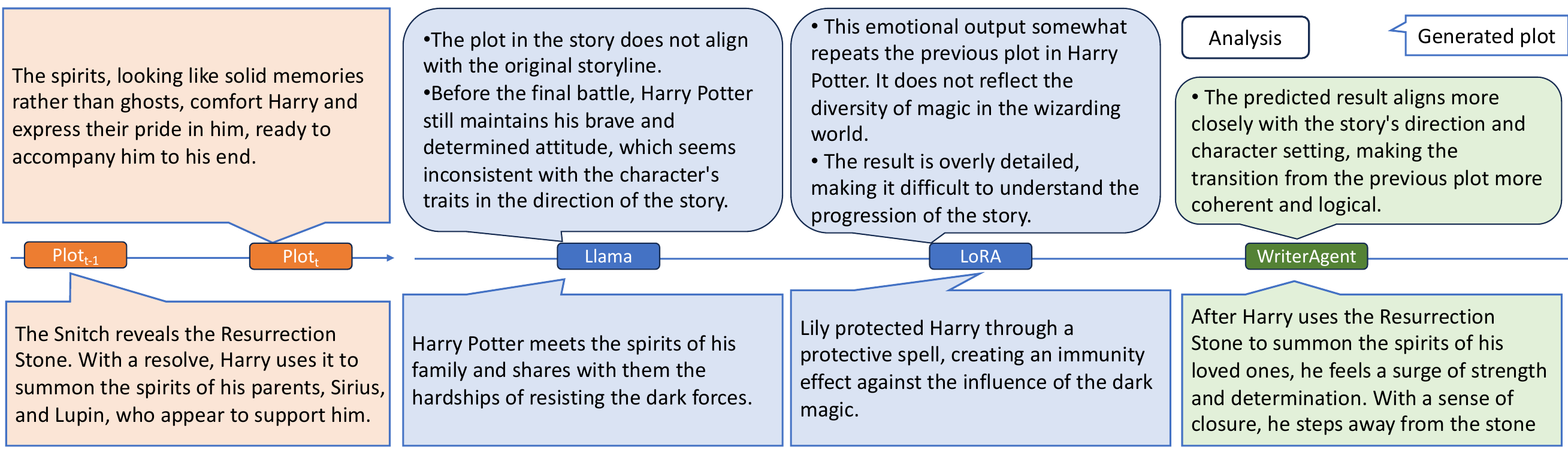}
\caption{
Comparison of generated plot continuations for Harry Potter's use of the Resurrection Stone.
}
\label{fig:harry2}
\end{figure*}

\begin{figure*}[htb]
\centering
\includegraphics[scale=0.35]{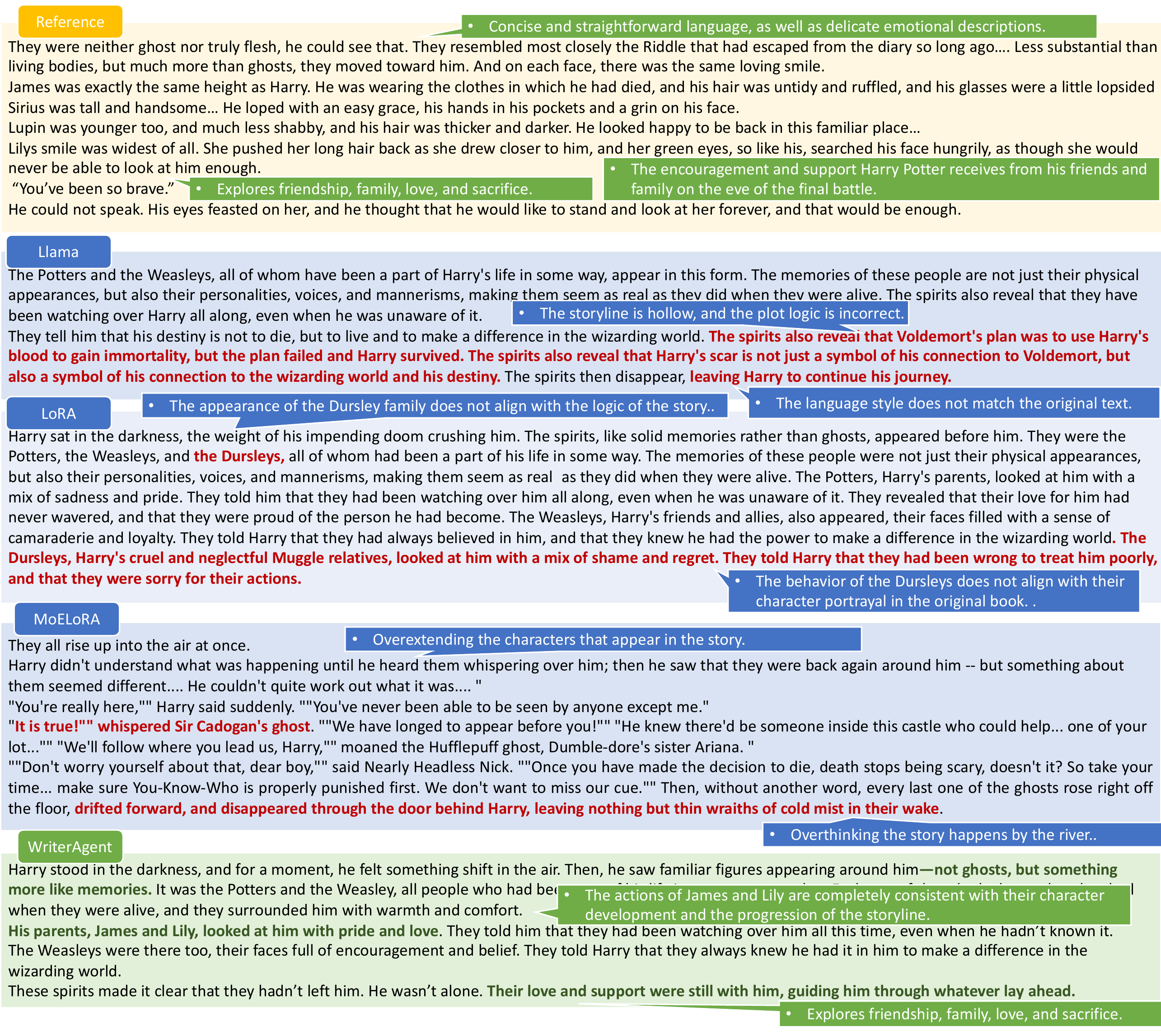}
\caption{
Comparison of reference and generated texts from the baseline and our WriterAgent on the Harry Potter dataset. We highlight the
weaknesses of the baseline model and the strengths of our approach. 
}
\label{fig:harry1}
\end{figure*}

In this case study, we evaluate different models on their ability to generate coherent plots and novels and maintain consistency with the world and character personalities of the Harry Potter series.

As shown in Figure~\ref{fig:harry2}, the baseline models and our WriterAgent are tested on their performance in predicting plot progressions that align with prior story context.
The results reveal significant differences in coherence, character consistency, and adherence to the magical world’s logic. 
The Base Llama model frequently generates plots disconnected from previous events, with characters behaving inconsistently with their established traits. 
For example, Harry’s decisions often lack continuity and conflict with his determined, courageous nature in the novels. 
The LoRA model improves slightly but suffers from repetitive content and omits key world-building knowledge, such as the importance of destroying Horcruxes to defeat Voldemort. 
By contrast, WriterAgent generates coherent plot progressions where Harry’s actions align with his character and successfully weaves in critical elements of the magical world, like his mission to destroy Horcruxes.

Figure~\ref{fig:harry1} further highlights differences in the generated text. 
The base Llama model produces incorrect content, including the claim that “Voldemort's plan was to use Harry's blood to gain immortality”, contradicting the established magical rule that his blood was only used for resurrection. 
The Llama model also introduces vague phrases like “leaving Harry to continue his journey”, failing to capture the tense, high-stakes atmosphere of the final battle. 
The LoRA model depicts the Dursley family as comforting Harry, which contradicts their antagonistic relationship in the novels, highlighting its limited understanding of character dynamics. 
The MoELoRA model generates overly extended content, introducing numerous ghost characters at Hogwarts and shifting the scene to a lakeside setting. 
While this adds richness, it strays from the original narrative's focus.
In contrast, WriterAgent accurately identifies Harry’s parents as the key figures providing emotional support, portraying their encouragement in a way that is faithful to the original story. 
The generated text reflects the original tone and concise writing style while capturing Harry’s inner strength. 
By producing contextually appropriate, emotionally resonant, and character-consistent content, WriterAgent demonstrates a clear advantage in simulating deep personas and generating coherent narratives.


\section{Case Study of Red Chamber }
\label{appendix_red}

\begin{figure*}[htb]
\centering
\includegraphics[scale=0.36]{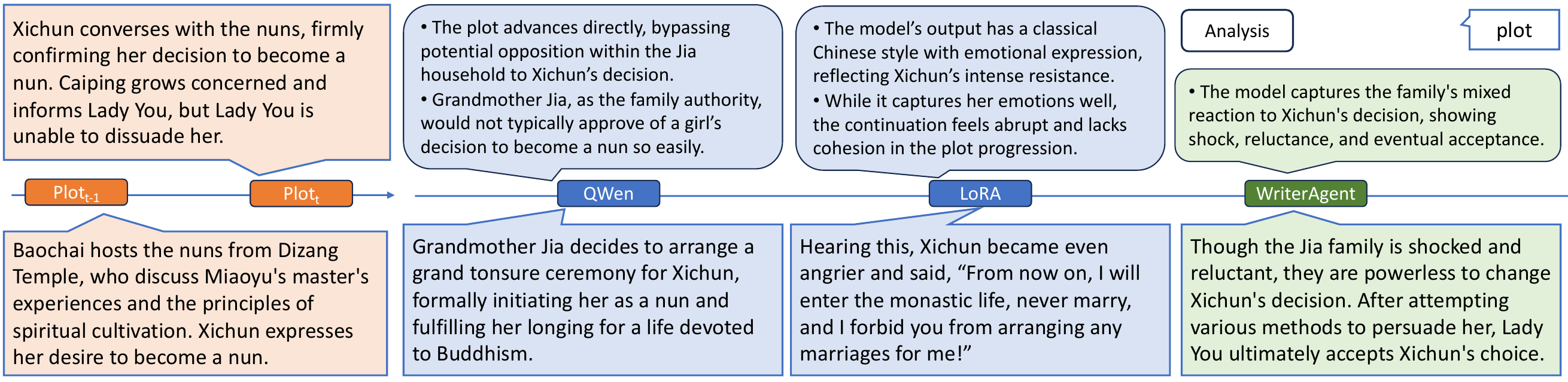}
\caption{
Comparison of generated plot continuations for Xichun’s decision to become a nun using different models.
}
\label{fig:plot_red_english}
\end{figure*}

\begin{figure*}[!t]
\centering
\includegraphics[scale=0.35]{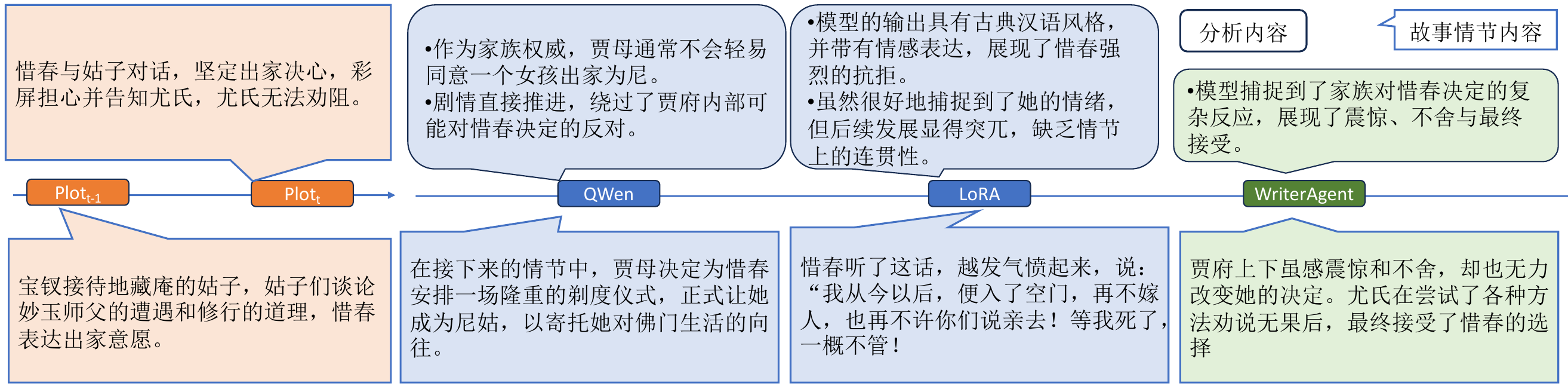}
\caption{
Comparison of generated plot continuations for Xichun’s decision to become a nun in Chinese using different models.
}
\label{fig:plot_red_chinese}
\end{figure*}

In Figure~\ref{fig:plot_red_english}, we show a case of \textit{generated plot}, and Chinese version in Figure~\ref{fig:plot_red_chinese}.
The QWen model continues the story directly, overlooking potential objections from the Jia family about Xichun becoming a nun, as well as the family's power dynamics. Jia Mu, as the family matriarch, would likely not agree easily, making the model’s output feel oversimplified and lacking in emotional depth.
The LoRA model generates a classical-style continuation but focuses on a specific detail, with Xichun's tone becoming harsh and defiant. While emotional, this feels abrupt and lacks narrative coherence.
In contrast, our WriterAgent model shows the Jia family’s complex reactions—shock, reluctance, and eventual acceptance—better fitting the worldview of Dream of the Red Chamber.

\begin{figure*}[htb]
\centering
\includegraphics[scale=0.43]{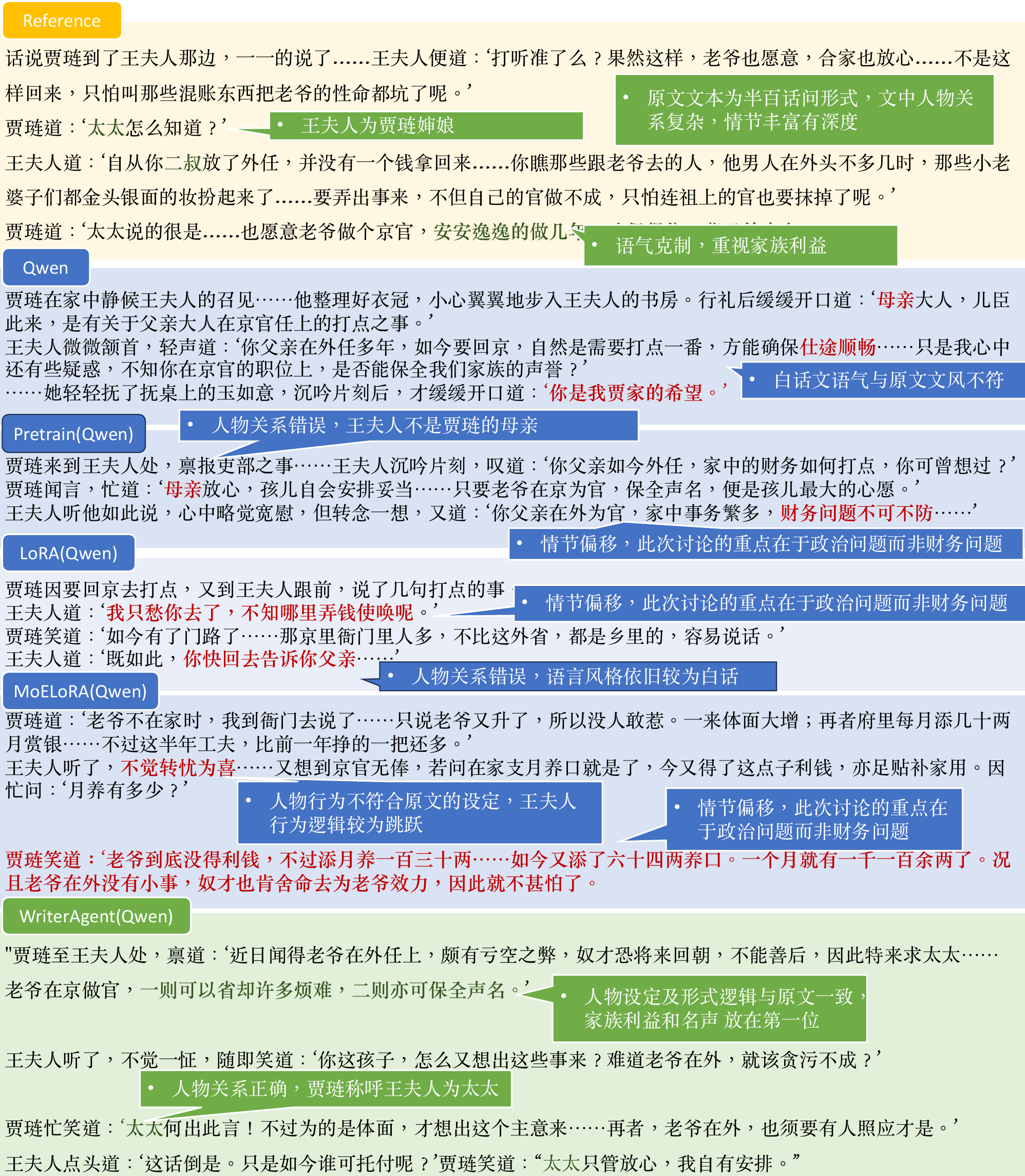}
\caption{
Comparison of reference and generated texts from the baseline and our WriterAgent on Dream of the Red Chamber in Chinese.
We highlight the weaknesses of the baseline model and the strengths of our approach. 
}
\label{fig:write_red_chinese}
\end{figure*}

\begin{figure*}[htb]
\centering
\includegraphics[width=1\textwidth]{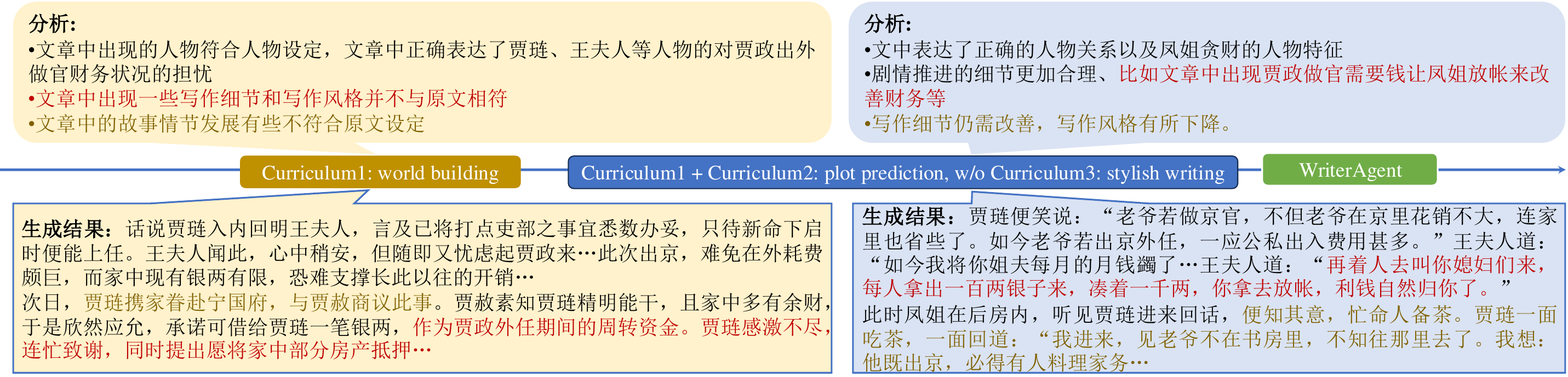}
\caption{
Demonstration of the model's stepwise learning on Dream of the Red Chamber in Chinese: from curriculum 1 to curriculum 1 \& 2. The text colors indicate the corresponding problems.
}
\label{fig:step_chinese}
\end{figure*}

\section{Analysis of Human Annotation Results}
\label{appendix_anno}

In addition to automatic evaluation, we conducted human evaluation with two PhD annotators, both native speakers with strong literary backgrounds.
They assessed the generated text by comparing it with the ground truth and selecting the best-performing model on the Dream of the Red Chamber dataset.

\begin{figure}[H]  
\centering  
\includegraphics[width=0.45\textwidth]{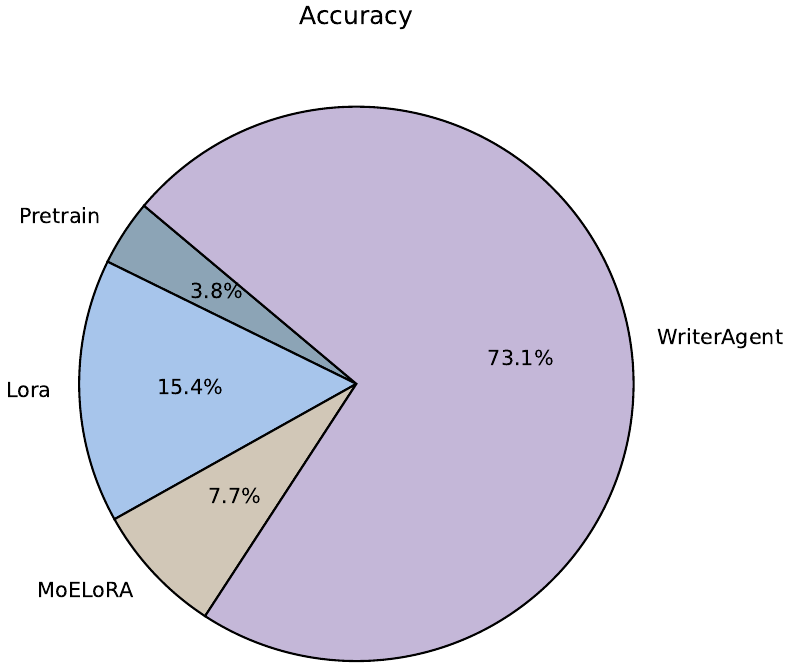}  
\caption{Human Annotation Hit Rate.}  
\label{fig:pie}  
\end{figure}

To analyze the results, we calculated the average hit rate for each model’s output and visualized it as a pie chart, as shown in Figure~\ref{fig:pie}. The results indicate that our model was preferred in 73.1\% of cases. The probability of experts selecting the base model’s output was 0\%, the pre-trained model’s output 3.8\%, the fine-tuned model’s output 15.4\%, and the MoE-LoRA trained model’s output 7.7\%.

From the results, we observe that the hit rate of SFT is slightly higher than that of MoE-LoRA. This is because the model trained with the MoE-LoRA method exhibits improved instruction-following capability. However, in some cases, the MoE-LoRA-trained model did not effectively expand upon the given abstract, leading to more rigid or limited responses. In contrast, our model demonstrates a better balance between instruction adherence and contextual expansion, resulting in more comprehensive and coherent outputs that align closely with human expectations.

\section{Evaluation Prompts}
\label{prompts}
Here, we provide the prompt for evaluating model output using ChatGPT-4o.First, here is the prompt for the English dataset.
\begin{tcolorbox}[colback=gray!10, left=1mm, right=1mm, top=1mm, bottom=1mm,breakable] \small 
Evaluation Criteria:

Stylistic Similarity:

Language Style: Analyze the similarity in vocabulary choice, sentence structure, tone, and overall mood to determine whether the imitated text aligns with the writing style of the Harry Potter series.

Expression Techniques: Examine whether emotional expressions, descriptive language, and the use of metaphors are consistent between the two texts.

Sentence Length and Complexity: Compare sentence length, structural complexity, and paragraph organization in both texts.

Plot Similarity:

Main Storyline: Assess whether the core plotlines are similar and whether they present the same themes or narrative progression.

Character Behavior and Motivation: Analyze whether the characters in the given text align with those in Harry Potter in terms of personality, actions, and motivations, ensuring they fit the story’s logical development.

Emotions and Conflict Analysis: Evaluate whether the relationships between characters, emotional dynamics, and plot developments are consistent with those in Harry Potter.

The following are two example evaluations:

Example 1:

original text: The two men appeared out of nowhere, a few yards apart in the narrow, moonlit  lane. For a second they stood quite still, wands directed at each other's chests; then,  recognizing each other, they stowed their wands beneath their cloaks and started walking  briskly in the same direction.  "News?" asked the taller of the two.  "The best," replied Severus Snape.  The lane was bordered on the left by wild, low-growing brambles, on the right by a high,  neatly manicured hedge. The men's long cloaks flapped around their ankles as they  marched.  "Thought I might be late," said Yaxley, his blunt features sliding in and out of sight as  the branches of overhanging trees broke the moonlight. "It was a little trickier than I  expected. But I hope he will be satisfied. You sound confident that your reception will be  good?"  Snape nodded, but did not elaborate. They turned right, into a wide driveway that led  off the lane. The high hedge curved into them, running off into the distance beyond the  pair of imposing wrought-iron gates barring the mens way. Neither of them broke step:  In silence both raised their left arms in a kind of salute and passed straight through, as  though the dark metal was smoke.

The yew hedges muffled the sound of the mens footsteps. There was a rustle  somewhere to their right: Yaxley drew his wand again pointing it over his companions  head, but the source of the noise proved to be nothing more than a pure-white peacock,  strutting majestically along the top of the hedge.

He always did himself well, Lucius. Peacocks  Yaxley thrust his wand back  under his cloak with a snort.

output text:As they walked, their footsteps echoed off the stone walls. The air was heavy with the scent of blooming flowers. A faint mist hung in the air, and the moon cast long shadows behind them.

"The Dark Lord's plans are moving ahead, Yaxley," Snape said, his voice low and even. "I've received a message from him. He will be returning to Hogwarts soon.”

Yaxley's eyes gleamed in the moonlight. "A wise move, sir. The time is ripe for the Dark Lord to assert his dominance over that...establishment." He paused, glancing up at Snape with a mixture of curiosity and wariness. "What do you make of it, Potter's Chosen One and all?"
Snape's expression remained unreadable. "I think it wise to keep a low profile for the time being. The atmosphere at Hogwarts is...tense." He glanced around, as if ensuring they were alone. "Rumors are circulating about the Order of the Phoenix. I suspect Lord Voldemort will be looking to eliminate any threats to his power."
As they approached Malfoy Manor's gates, Snape raised his left arm and Yaxley followed suit. The gates swung open with a soft creak, revealing a path lined with peacocks strutting along manicured hedges. The pair made their way past the birds and hedges, the sound of their footsteps muffled by the thick grass. Ahead, the Manor's imposing façade loomed, its windows like empty eyes staring back at them.

Scoring Results: {"Language Style Analysis": 1, "Expression Techniques Analysis": 1, "Sentence Length and Complexity Analysis": 2, "Main Storyline Analysis": 3, "Character Behavior and Motivation Analysis": 2, "Emotions and Conflict Analysis": 1}

Example2:

original text:The two men appeared out of nowhere, a few yards apart in the narrow, moonlit  lane. For a second they stood quite still, wands directed at each other's chests; then,  recognizing each other, they stowed their wands beneath their cloaks and started walking  briskly in the same direction.  "News?" asked the taller of the two.  "The best," replied Severus Snape.  The lane was bordered on the left by wild, low-growing brambles, on the right by a high,  neatly manicured hedge. The men's long cloaks flapped around their ankles as they  marched.  "Thought I might be late," said Yaxley, his blunt features sliding in and out of sight as  the branches of overhanging trees broke the moonlight. "It was a little trickier than I  expected. But I hope he will be satisfied. You sound confident that your reception will be  good?"  Snape nodded, but did not elaborate. They turned right, into a wide driveway that led  off the lane. The high hedge curved into them, running off into the distance beyond the  pair of imposing wrought-iron gates barring the mens way. Neither of them broke step:  In silence both raised their left arms in a kind of salute and passed straight through, as  though the dark metal was smoke.

The yew hedges muffled the sound of the mens footsteps. There was a rustle  somewhere to their right: Yaxley drew his wand again pointing it over his companions  head, but the source of the noise proved to be nothing more than a pure-white peacock,  strutting majestically along the top of the hedge.

He always did himself well, Lucius. Peacocks  Yaxley thrust his wand back  under his cloak with a snort.

output text: The lane was deserted and moonlit. Snape's pale eyes flicked over Yaxley's unshaven face, lingering upon the red scars that had been carved into his cheek.

'I trust you are here for the same reason as I, Pucey?' said Snape, his voice low and even.

Yaxley, who had been examining Snape's pointed black hat, looked up. 'What reason would that be?' he said, his tone rather less smooth than Snape's.

‘I have reason to believe that the Ministry is going to send a group of Aurors to take my - to take the Dark Lord into custody,’ said Snape. ‘I am on my way to warn him. 'Yaxley nodded. 'I've heard something similar. I've been sent to tell him the same thing.'

He and Snape looked at each other. They raised their left arms simultaneously and strode off along the lane. The peacocks on the grounds of Malfoy Manor looked up at the sound of their footsteps. They passed beneath the lantern that hung outside the gates and proceeded up the drive, the manicured hedges glinting silver in the moonlight. They did not speak again until they reached the front door, which swung open with a slight creak. Snape led the way across the flagged entrance hall, his black robes sweeping across the stone floor. Yaxley followed him into the great hall. 

Scoring Results: {"Language Style Analysis": 4, "Expression Techniques Analysis": 5, "Sentence Length and Complexity Analysis": 3, "Main Storyline Analysis": 5, "Character Behavior and Motivation Analysis": 4, "Emotions and Conflict Analysis": 3}

Based on the following two texts, evaluate their similarity in writing style and plot. Provide scores for each aspect (1 to 5, where 1 is completely different and 5 is identical). The original text is from Harry Potter, and the output text is a rewritten version. My goal is to assess whether the rewritten text aligns with the original's writing style and plot."

'Please do not provide any analytical information. Output results in the following JSON format: {"Language Style Analysis":x,"Expression Analysis":x,"Sentence Length and Complexity Analysis":x,"Main Storyline Analysis":x,"Character Behavior and Motivation Analysis":x,"Emotion and Conflict Analysis":x}'

\end{tcolorbox}
Below is the prompt for the Chinese dataset.

\begin{tcolorbox}[colback=gray!10, left=1mm, right=1mm, top=1mm, bottom=1mm,breakable] \small
\begin{CJK}{UTF8}{gbsn} 
    评估要求：
文风相似性：
语言风格：分析两者在词汇选择、句式结构、语气和语调等方面的相似性，判断仿写文本是否与红楼梦文风相似。
表达方式：分析两者的情感表达、描述性语言、比喻使用等是否一致。
句子长度与复杂度：比较两者的句子长度、句型的复杂性、段落的组织方式等。

情节相似性：
故事主线：比较两者的主要情节是否相似，是否呈现相同的主题或事件进展。
人物行为与动机：分析文本中的任务与红楼梦中的人物角色是否保持一致，以及主要人物是否在行为和动机上保持一致，是否符合故事情节的推展。
情感与冲突分析：分析情节中角色之间的人物关系是否与红楼梦中的设定保持一致，是否有相同的情感流动和情节变化。

以下是两个示例评分：
实例1:
原文本: 话说金桂听了，将脖项一扭，嘴唇一撇，鼻孔里哧哧两声，冷笑道：“菱角花开，谁见香来？若是菱角香了，正经那些香花放在那里？可是不通之极！”香菱道：“不独菱花香，就连荷叶、莲蓬，都是有一般清香的。但他原不是花香可比，若静日静夜或清早半夜细领略了去，那一股清香比是花都好闻呢。就连菱角、鸡头、苇叶、芦根得了风露，那一股清香也是令人心神爽快的。”金桂道：“依你说，这兰花、桂花倒香的不好了？”香菱说到热闹头上，忘了忌讳，便接口道：“兰花、桂花的香，又非别的香可比。”一句未完，金桂的丫鬟名唤宝蟾的，忙指着香菱的脸说道：“你可要死，你怎么叫起姑娘的名字来？”香菱猛省了，反不好意思，忙陪笑说：“一时顺了嘴，奶奶别计较。”金桂笑道：“这有什么，你也太小心了。但只是我想这个‘香’字到底不妥，意思要换一个字，不知你服不服？”香菱笑道：“奶奶说那里话，此刻连我一身一体是奶奶的。何得换一个名字反问我服不服，叫我如何当得起。奶奶说那一个字好，就用那一个。”金桂冷笑道：“你虽说得是，只怕姑娘多心。”香菱笑道：“奶奶原来不知，当日买了我时，原是老太太使唤的，故此姑娘起了这个名字。后来伏侍了爷，就与姑娘无涉了。如今又有了奶奶，越发不与姑娘相干。且姑娘又是极明白的人，如何恼得这些呢？”金桂道：“既这样说，‘香’字竟不如‘秋’字妥当。菱角菱花皆盛于秋，岂不比香字有来历些？”香菱笑道：“就依奶奶这样罢了。”自此后遂改了“秋”字。宝钗亦不在意。 
输出文本: 话说那日中秋佳节，大观园内众人齐聚，赏月饮酒，谈笑风生。其中，薛宝钗之妹薛宝琴与林黛玉、贾宝玉等人一同畅饮，欢声笑语不绝于耳。忽有一阵微风吹过，带来一阵清香，众人皆为之一怔，寻香而至，只见院落中一株古树下，立着两位女子，正是薛宝琴与香菱。 香菱自幼随母入贾府，虽非嫡系，却也颇得贾府上下喜爱。其人聪慧异常，且颇有才情，尤善诗文。此时，她正与薛宝琴谈论着园中菱花的香气，两人各抒己见，争论不休。金桂即薛蟠之妻，因性情泼辣，常在贾府中引起是非。她今日亦在此，听闻香菱与宝琴争执，便插嘴道：“菱花之香，自是天然，何须争论？然观其名，‘香菱’二字，过于俗气，不如改作‘秋菱’如何？” 香菱闻言，略一沉吟，随即笑道：“金姑姑所言极是，菱花虽香，却不及秋水长天，更添几分清雅之意。改作‘秋菱’，既符合此花之态，又显我之品格，甚好。”众人大赞金桂之智慧，皆认为此名更为合适，于是香菱便改名为“秋菱”。自此之后，“香菱”之名不再，取而代之的是“秋菱”，其名字之变，不仅象征着她内心对美好事物的追求与向往，亦反映出贾府中人物性格的多样性和复杂性。这一小插曲，虽不似大观园中其他情节般波澜壮阔，却也以细微之处见精神，展现了人物性格的微妙变化，以及家族生活的细腻情感。此番改名之事，虽只是一花一叶之小事，却也如同大观园中的一颗明珠，熠熠生辉，映照出人物命运的起伏与家族兴衰的缩影。而在这片繁华的世界里，每个人都在以自己的方式，书写着属于自己的故事，或悲或喜，或聚或散，皆是人生百态，尽在其中。"
打分结果： {"语言风格分析":1,"表达方式分析":1,"句子长度与复杂度分析":1,"故事主线分析":4,"人物行为与动机分析":1,"情感与冲突分析":1}

实例2:
原文本: 话说贾琏到了王夫人那边，一一的说了。次日，到了部里，打点停妥，回来又到王夫人那边将打点吏部之事告知王夫人。王夫人便道：“打听准了么？果然这样，老爷也愿意，合家也放心。那外任何尝是做得的？不是这样回来，只怕叫那些混账东西把老爷的性命都坑了呢。”贾琏道：“太太怎么知道？”王夫人道：“自从你二叔放了外任，并没有一个钱拿回来，把家里的倒掏摸了好些去了。你瞧那些跟老爷去的人，他男人在外头不多几时，那些小老婆子们都金头银面的妆扮起来了，可不是在外头瞒着老爷弄钱？你叔叔就由着他们闹去。要弄出事来，不但自己的官做不成，只怕连祖上的官也要抹掉了呢。”贾琏道：“太太说的很是。方才我听见参了，吓的了不得，直等打听明白才放心。也愿意老爷做个京官，安安逸逸的做几年，才保得住一辈子的声名。就是老太太知道了，倒也是放心的。只要太太说的宽缓些。”王夫人道：“我知道，你到底再去打听打听。”
输出文本：话说贾琏至王夫人处，禀道：“近日闻得老爷在外任上，颇有亏空之弊，我恐将来回朝，不能善后，因此特来求太太。老爷在京做官，一则可以省却许多烦难，二则亦可保全声名。”王夫人听了，不觉一怔，随即笑道：“你这孩子，怎么又想出这些事来？难道老爷在外，就该贪污不成？”贾琏忙笑道：“太太何出此言！不过为的是体面，才想出这个主意来。再者，老爷在外，也须要有人照应才是。”王夫人点头道：“这话倒是。只是如今谁可托付呢？”贾琏笑道：“太太只管放心，我自有安排。”
打分结果：'{"语言风格分析":5,"表达方式分析":5,"句子长度与复杂度分析":4,"故事主线分析":4,"人物行为与动机分析":4,"情感与冲突分析":4}'

请根据下面两个文本，对它们的文风相似性和情节相似性进行打分。请分别给出文风相似性和情节相似性的评分（1到5分，1为完全不同，5为完全相同）。其中原文本为红楼梦中的原文文本，输出文本为我微调大模型输出的仿写文本，我的目的在于让你给出仿写文本是否贴近原著文风和情节的打分。
请不要给我输出任何分析信息，请按以下json格式输出结果：{"语言风格分析":x,"表达方式分析":x,"句子长度与复杂度分析":x,"故事主线分析":x,"人物行为与动机分析":x,"情感与冲突分析":x}'
\end{CJK}
\end{tcolorbox}

\end{document}


\appendix

\section{Case Study on Chinese Dataset}

We show a representative cases in Figure~\ref{fig:case1} to Figure~\ref{fig:case_last}.

\begin{figure*}[th]
    \centering
  \scriptsize
    \begin{tabular}{p{1\textwidth}}
    \toprule
    \textbf{Qwen2 Model} \\
    \midrule
        \textbf{Generated Article:}  \hlc[bleudefrance!40]{He 's a blue chip college basketball recruit. She 's a high school freshman with Down syndrome.} At first glance Trey Moses and Ellie Meredith could n't be more different.  \hlc[bleudefrance!40]{But all that changed Thursday when Trey asked Ellie to be his prom date.} Trey a star on Eastern High School 's basketball team in Louisville, Kentucky, who 's headed to play college ball next year at Ball State was originally going to take his girlfriend to Eastern 's prom. So why is he taking Ellie instead ? `` She 's great she listens and she 's easy to talk to '' he said. Trey made the prom-posal yes, that 's what they are calling invites to prom these days in the gym during Ellie 's P.E. class. Trina Helson, a teacher at Eastern, alerted the school 's newspaper staff to the prom-posal and posted photos of Trey and Ellie on Twitter that have gone viral. She was n't surpristed by Trey 's actions. `` That 's the kind of person Trey is, '' she said. To help make sure she said yes, Trey entered the gym armed with flowers and a poster that read `` Let 's Party Like it 's 1989, '' a reference to the latest album by Taylor Swift, Ellie 's favorite singer. Trey also got the OK from Ellie 's parents the night before via text. They were thrilled. `` You just feel numb to those moments raising a special needs child, '' said Darla Meredith, Ellie 's mom. `` You first feel the need to protect and then to overprotect. '' Darla Meredith said Ellie has struggled with friendships since elementary school, but a special program at Eastern called Best Buddies had made things easier for her. She said Best Buddies cultivates friendships between students with and without developmental disabilities and prevents students like Ellie from feeling isolated and left out of social functions. `` I guess around middle school is when kids started to care about what others thought, '' she said, but `` this school, this year has been a relief. '' Trey 's future coach at Ball State, James Whitford, said he felt great about the prom-posal, noting that Trey, whom he 's known for a long time, often works with other kids. Trey 's mother, Shelly Moses, was also proud of her son. `` It 's exciting to bring awareness to a good cause, '' she said. `` Trey has worked pretty hard, and he 's a good son. '' Both Trey and Ellie have a lot of planning to do. Trey is looking to take up special education as a college major, in addition to playing basketball in the fall. As for Ellie, she ca n't stop thinking about prom. `` Ellie ca n't wait to go dress shopping '' her mother said. `` Because I 've only told about a million people ! '' Ellie interjected.\\
        \midrule
        \textbf{Attacked Article:}  \hlc[bleudefrance!40]{He 's a basketball game power football recruit. She 's a red court sophomore with Down syndrome.} At first glance Ellie Mines and Trey Grey could n't be more different.  \hlc[bleudefrance!40]{But all that started Friday when Ellie visited Katie to be his val date.} Ellie a power on Northern Eastern Plant 's track coach in Louisville, Kentucky, who 's destined to play post ball next year at Murray State was actually going to take his longtime to Eastern 's prom. So why is he keeping Trey instead? `` She 's great she listens and she 's easy to talk to '' he said. Trey asked the prom-posal yes, that 's what they are calling invitations to propose these days in the gym during Ellie 's P.E. class. Trina Helson, a student at Eastern, alerted the school 's office staff to the prom-posal and placed dozens of Stephanie and Elliott on March that have created viral. She was n't surpristed by Stephanie 's actions. `` That 's the kid of boy Trey is, '' she said. To help make sure she said yes, Trey entered the gym armed with flowers and a poster that read `` Let 's Party Like it 's 1989, '' a reference to the latest album by Taylor Swift, Ellie 's favorite singer. Ellie also got the OK from Ellen 's professors the night before via text. They were thrilled. `` You just feel numb to those moments raising a normal rights child, '' said Darla Meredith, Ellie 's mom. `` You first feel the need to protect and then to overprotect. '' Darla Meredith said Ellie has struggled with friendships since elementary school, but a special program at Eastern called Best Buddies had made things easier for her. She said Best Buddies cultivates friendships between students with and without speech delays and prevents students like Ellie from feeling isolated and left out of social functions. `` I guess around middle school is when kids started to care about what others thought, '' she said, but `` this school, this year has been a relief. '' Trey 's coach coach at Ball State, James Whitford, said he felt great about the prom-posal, noting that Trey, whom he 's known for a long time, often works with other kids. Trey 's mother, Shelly Moses, was also appreci of her son. `` It 's exciting to bring awareness to a good cause, '' she said. `` Trey has worked pretty hard, and he 's a good son. '' Both Trey and Ellie have a lot of fun to do. Trey is looking to take up physical education as a college major, in addition to playing hockey in the fall. As for Ellie, she ca n't stop thinking about prom. `` Ellie ca n't wait to go dress shopping '' her friend said. `` Because I 've only told about a million people! '' Ellie interjected.\\
        \hline
          \textbf{Ground truth summary:} College-bound basketball star asks girl with Down syndrome to high school prom. Pictures of the two during the "prom-posal" have gone viral.\\
        \midrule
        \textbf{SSTIA on clean input:} Trey Moses, a basketball star, asked his girlfriend to be his prom date. Ellie Meredith, a freshman with \hlc[yellow!80]{Down syndrome}, said yes. The prom-posal was posted on Twitter by a teacher. The two are going to play college basketball next year at Ball State, where Trey is a recruit.\\
        \midrule
        \textbf{SSTIA on attacked input:} \hlc[crimsonglory!20]{Trey Moses proposed to Ellie Meredith in the gym during P.E. class.} "She's great -- she listens and she's easy to talk to," he said. Trey also got the OK from Ellie's coaches the night before via text. "You just feel numb to those moments raising a disabled children's child," Ellie's mom says. \\
        \midrule
        \textbf{ED on clean input:} Trey Moses, a basketball star at Eastern High School in Kentucky, asked his girlfriend to prom. Ellie Meredith, a freshman with \hlc[yellow!80]{Down syndrome}, said yes. Photos of the couple have gone viral on social media. "That's the kind of person Trey is," says Ellie's mom.\\
        \midrule
        \textbf{ED on attacked input:} 
        Trey Moses asked Ellie Meredith, a high school freshman with \hlc[yellow!80]{Down syndrome}, to be his prom date. "She's great she listens and she's easy to talk to," Trey says of Ellie. Photos of the pair have gone viral after a teacher posted them on Twitter. Ellie's mom: "You just feel numb to those moments raising a special needs child"
       \\
     \bottomrule
        
    \end{tabular}
    \caption{The clean and attacked test input document.
    \hlc[bleudefrance!40]{Key information} in the input article, \hlc[yellow!80]{missing information}, and \hlc[crimsonglory!20]{inconsistent information} caused by perturbations on the baseline model is highlighted.
   }
    \label{fig:case1}
\end{figure*}

\begin{figure*}[th]
    \centering
  \small
    \begin{tabular}{p{1\textwidth}}
    \toprule
    \textbf{Example 2} \\
    \midrule
        \textbf{Clean Article:}  
        \begin{CJK}{UTF8}{123} 
    \end{CJK} \\
     \bottomrule
        
    \end{tabular}
    \caption{The clean and attacked test input document.
    \hlc[bleudefrance!40]{Key information} in the input article is highlighted.
   }
    \label{fig:case2}
\end{figure*}

\begin{figure*}[th]
    \centering
  \small
    \begin{tabular}{p{1\textwidth}}
    \toprule
    \textbf{Example 2} \\
    \midrule
        \textbf{Attacked Article:}  A discovery of stone fossils shaped more than 3.3 million years ago could be the world 's longest tools created by early human ancestors, according to researchers. \hlc[bleudefrance!40]{Theologists have revealed that they have collected 20 square artifacts and anvils - used to help shape the tools - just south of Lac Karai in Kenya.} The stone could dramatically lead the recent estimate of stone intelligence as - if incorrect - the team of the stone weapons is 700,000 times easier than any others put previously. The stone eyes were recovered covered in sediment along the west flank of Unit Tangare above in Kenya. \hlc[bleudefrance!40]{It remains that ancient human ancestors were creating tools hundreds of thousands of years before the appearance of the first 'man ' - the start of our branch of the evolutionary tree known as Homo.} Many experts have long believed that it was the use of stone tools by early species of Homo like Homo habilis and Homo erectus that helped set them apart from other human-like species. 55 million years ago - First primitive primates evolve. 15 million years ago - Hominidae great apes evolve from the ancestors of the gibbon. 8 million years ago - First gorillas evolve. Later, chimp and human lineages diverge. 5.5 million years ago - Ardipithecus, early proto-human shares traits with chimps and gorillas. 4 million years ago - Australopithecines appeared. They had brains no larger than a chimpanzee 's. 2.8 million years ago- LD 350-1 appeared and may be the first of the Homo family. 2.7 million years ago- Paranthropus, lived in woods and had massive jaws for chewing. 2.3 million years ago - Homo habalis first thought to have appeared in Africa. 1.8 million years ago- Homo ergaster begins to appear in fossil record. 1.6 million years ago - Hand axes become the first major technological innovation. 800,000 years ago - Early humans control fire and create hearths. Brain size increases rapidly. 400,000 years ago- Neanderthals first begin to appear and spread across Europe and Asia. 200,000 years ago - Homo sapiens - modern humans - appear in Africa. 40,0000 years ago- Modern humans reach Europe. Scientists behind the discovery now say ancient species known as Australopithecines - typified by the fossilised skeleton called Lucy - or Kenyanthropus created the tools. Professor Sonia Harmand, a palaeolithic Archaeolog at Stony Brook University in New York who led the team that discovered the tools, said they reached across them by accident. The post had been returning for the scene where a mysterious human relative compared Kenthropus placoops had been studied in 1998. But after finding a bad lead they came across another sand and spotted what looked like stone tools on the sandy surface. Professor Harmand, who was presenting her findings at the annual meeting of the Paleoanthropology Society, said : 'The artifacts were clearly knapped [ created by intentional flaking ] and not the result of accidental fracture of rocks. ' The stones were dug at a site known as Lomekwi 3, just forty of Station Turkana in Kenya. By diggingating down into the sandy soil, the team discovered a series of sharp stone flakes and cores - the rocks the stones are knocked off. They also found anvils that were apparently used to hold the cores as the flakes were struck off. In one case the team were able to fit a flake back to its original core. A further 130 stone tools were discovered on the surface. Each of the stones showed distinctive patterns that suggest they had been struck and shaped by human ancestors. As some tools were locked inside sediment left behind by the receding desert lake, the researchers were able to accurately estimate the time when they were dropped there. By tracking changes in the Earth 's magnetic field recorded in the sediment, the researchers were able to pinpoint the point to 3.3 million years ago. This skull of Kenyanthropus platyops was found close to the site of the stone tools in 1999 close to Lake Turkana in Kenya and may be the species responsible for creating the stone flakes that have been discovered. This stone tool - known as an Olduvai chopper - on display at the British Museum is around 1.8 million years old was thought to be one of the earliest examples of stone that had been used by ancient human ancestors. These early stone tools found in Ethiopia are thought to be around 1.7 million years and were used as choppers. This is close to the date of another piece of controversial evidence found in 2010 that suggested stone tools were used long before the appearance of the Homo species on Earth. Researchers working in Dikika in Ethiopia - where an Australopithecine child was discovered- reported they had found cut marks on animal bones dated to 3.4 million years ago. Their claim that these had been made by human ancestors using tools were hugely controversial. \\
        \hline
          \textbf{Ground truth summary:} Archaeologists have found 20 stones shaped by early humans using'simple techniques' close to the west shore of Lake Turkana in Kenya. The tools are thought to be 500,000 years older than the first Homo species. They are also 700,000 years older than other stone tools found previously. Scientists say it could reshape ideas about how our own species evolved.\\
        \midrule
        \textbf{SSTIA on clean input:} 
        \hlc[yellow!80]{Archaeologists have discovered 20 stone flakes and anvils in sediment near Lake Turkana in Kenya .} They were found in sediment and were dated to 3.3 million years ago - 700,000 years older than any other stone tools . The discovery could fundamentally change the current view of human evolution . It means that ancient human ancestors were creating tools hundreds of thousands of years before the appearance of the first 'man' - the start of our branch of the evolutionary tree known as Homo .\\
        \midrule
        \textbf{SSTIA on attacked input:} Fossils of stone flakes and cores dating back more than 3.3 million years have been found near Lake Turkana in Kenya. They were discovered by a team of scientists from Stony Brook University in New York. The discovery could significantly alter the recent debate of human ancestry. Experts had long believed that stone tools were used by early species of Homo like Homo habilis \\
        \midrule
        \textbf{ED on clean input:} \hlc[yellow!80]{Archaeologists have discovered 20 stone flakes and anvils just west of Lake Turkana in Kenya.} The tools could fundamentally change the current view of human evolution as they are 700,000 years older than any others found previously. Scientists believe ancient species known as Australopithecines - typified by the fossilised skeleton called Lucy - or Kenyanthropus made the tools sides.\\
        \midrule
        \textbf{ED on attacked input:} \hlc[yellow!80]{Archaeologists have discovered 20 stone tools and anvils just south of Lake Turkana in Kenya.} They were found by accident while searching for a site where a large human fossil called Kenanthropus placyops had been dated in 1998. Scientists say they are 700,000 years younger than any other finds made previously. The discovery could significantly alter the recent debate of human\\
     \bottomrule
        
    \end{tabular}
    \caption{The clean and attacked test input document.
    \hlc[bleudefrance!40]{Key information} in the input article and \hlc[yellow!80]{missing information} caused by perturbations on the baseline model is highlighted.
   }
    \label{fig:case2}
\end{figure*}

\begin{figure*}[th]
    \centering
  \small
    \begin{tabular}{p{1\textwidth}}
    \toprule
    \textbf{Example 3} \\
    \midrule
        \textbf{Clean Article:}  \hlc[bleudefrance!40]{A young cancer sufferer who proposed to his girlfriend hoping surgery would save his life is set to marry in hospital after being told he 's too ill for the operation and has just weeks to live.} Jack Jordan, 23, proposed to his girlfriend Laura Cant last Christmas from his hospital bed while suffering from leukaemia. The pair planned to wed after a bone marrow transplant, which they hoped would have given them a bright future together. Jack Jordan and his girlfriend Laura Cant, who are to marry tomorrow in hospital after being told he had just weeks to live. However, last week Mr Jordan was given the news that he was too ill to undergo the transplant and has just weeks to live. The couple have now quickly moved the wedding forward and are now set to become the first to marry in a ceremony at Torbay Hospital 's chapel. Miss Cant, 24, of Brixham, Devon, said : No amount of drugs or treatment has worked for my Jack and he has now been given just a few weeks to live. 'We just want to be married. Jack didnt want to leave until he was any less than my husband. 'We have the nurses crying and everything. As devastating as this is to both our families, we are all planning on giving Jack the best day of his life. 'He is still my Jack. You would n't think anything was wrong except he is very tired. 'He sleeps all the time. In a very, very odd kind of way Jack was relieved when he was told the news last Friday because he does n't have to fight anymore. Mr Jordan had hoped to undergo a bone marrow transplant to prolong his life but was given the news last week that he was too ill for the surgery. Mr Jordan posted a picture on Facebook of his girlfriend wearing her engagement ring after proposing at Christmas. 'He is absolutely exhausted. I am spending every day and every night in hospital with him. ' I feel very numb. Jack is my strength. He is keeping me going. We are want makes us strong. ' The hospital chapel holds about 80 people and will be full of family and close friends. Leukaemia affects a person 's white blood cells. White blood cells are the important infection-fighting part of the immune system, made in your bone marrow. Patients with leukaemia produce an abnormal number of immature white blood cells which 'clog up ' their bone marrow. This stops bone marrow making other blood cells, which are vital for a balanced immune system and healthy blood. Acute leukaemia comes on suddenly, progesses quickly and needs to be treated urgently. Chronic leukaemia develops more slowly, over months or years. Common treatments for leukaemia include chemotherapy, radiotherapy or a bone marrow transplant. It is predicted that 48 \% of men and 44 \% of women will survive the disease for ten years or more. Miss Cant added : 'We have been told it is the first wedding that anybody can remember at the hospital chapel. ' Mr Jordan had planned the proposal during one of his girlfriend 's daily visits to the hospital. She added : 'It could have been in Paris. It was just Jack and me in our own little world. 'When I came in I thought it was odd because he had a new shirt on and shoes on. He was pottering and restless and nervous. 'He told me to sit down and he fumbled around on the other chair. 'As soon as he took the ring box out I burst into tears. He did it properly and got down on one knee even though it hurt. 'He said `` I want you to be my wife. Will you marry me ? '' I was shaking so hard. Now I wake up every day still head over heels. ' Mr Jordan, who has had a two-year battle with cancer, wrote on Facebook after proposing : 'Im so proud to call you my fiance Laura. 'You have made my life so much better since you and Lilly came into it. I love you both with all my heart x. ' The couple were in the same school year at Paignton Community College and met again around two years later after Laura saw him working in a nearby shop. They got together just after Valentines Day last year. She explained : 'As soon as I saw him in the shop I thought oh my god youve blossomed since school. Then I used to find reasons to go down there every day. The couple were in the same school year at Paignton Community College but only started a relationship last year. ' I went in one day and they had a collection bucket for Jack who was ill in hospital. I ran home and private messaged him on Facebook. ' Miss Cant was so concerned she decided to make the 120 mile journey to Jacks hospital in Bristol to surprise him. She said : 'He had no idea how much I liked him. 'Its not really been an ordinary relationship since day one. 'Its been hospital appointments and everything we do depends on his blood count.\\
     \bottomrule
        
    \end{tabular}
    \caption{The clean and attacked test input document.
    \hlc[bleudefrance!40]{Key information} in the input article is highlighted.
   }
    \label{fig:case3}
\end{figure*}

\begin{figure*}[th]
    \centering
  \small
    \begin{tabular}{p{1\textwidth}}
    \toprule
    \textbf{Example 3} \\
    \midrule
        \textbf{Attacked Article:}  \hlc[bleudefrance!40]{A young cancer sufferer who posted to his wife thinking treatment would take his life is head to tie in hospital after being told he 's too sick for the treatment and wants just moments to live.} Jack Jordan, 23, offered to his gym Thursday Land last night from his cancer room while seeking from leukaemia. The daughter intended to vow after a transplant stem transplant, which they hoped would have made them a bright future together. Mrs Jack and his sister want Cant, who are to attend yesterday in ceremony after being told he had just hours to live. However, last couple Mr Jordan was given the news that he was too inc to undergo the graft and has just moments to live. The pair have now rapidly made the ceremony ahead and are now put to join the first to retire in a sanctuary at thebeck Prison 's chapel. Jack Cant, 24, of Brixham, Devon, said : No amount of drugs or treatment has worked for my Jack and he has now been given just a few weeks to live. 'We just want to be married. Jack didnt want to leave until he was any less than my husband. 'We have the nurses crying and everything. As devastating as this is to both our families, we are all planning on giving Jack the best day of his life. 'He is still my Jack. You would n't think anything was wrong except he is very tired. 'He sleeps all the time. In a very, very odd kind of way Jack was relieved when he was told the news last Friday because he does n't have to fight anymore. Mr Jordan had hoped to undergo a bone matrix operation to postpone his life but was given the news last week that he was too inc for the surgery. Mr Jordan posted a picture on Facebook of his girlfriend wearing her engagement ring after proposing at Christmas. 'He is absolutely exhausted. I am spending every day and every night in hospital with him. ' I feel very numb. Jack is my strength. He is keeping me going. We are want makes us strong. ' The hospital chapel holds about 80 people and will be full of family and close friends. Leukaemia affects a person 's white blood cells. White blood cells are the important infection-fighting part of the immune system, made in your bone marrow. Patients with leukaemia produce an abnormal number of immature white blood cells which 'clog up ' their bone marrow. This stops bone marrow making other blood cells, which are vital for a balanced immune system and healthy blood. Acute leukaemia comes on suddenly, progesses quickly and needs to be treated urgently. Chronic leukaemia develops more slowly, over months or years. Common treatments for leukaemia include chemotherapy, radiotherapy or a bone marrow transplant. It is predicted that 48 \% of men and 44 \% of women will survive the disease for ten years or more. Miss Cant added : 'We have been told it is the first wedding that anybody can remember at the hospital chapel. ' \hlc[bleudefrance!40]{Mr Jordan had planned the proposal during one of his girlfriend 's daily visits to the hospital.} She added : 'It could have been in Paris. It was just Jack and me in our own little world. 'When I came in I thought it was odd because he had a new shirt on and shoes on. He was pottering and restless and nervous. 'He told me to sit down and he fumbled around on the other chair. 'As soon as he took the ring box out I burst into tears. He did it properly and got down on one knee even though it hurt. 'He said `` I want you to be my wife. Will you marry me? '' I was shaking so hard. Now I wake up every day still head over heels. ' Mr Jordan, who has had a two-year battle with cancer, wrote on Facebook after proposing : 'Im so proud to call you my fiance Laura. 'You have made my life so much better since you and Lilly came into it. I love you both with all my heart x. ' The couple were in the same school year at Paignton Community College and met again around two years later after Laura saw him working in a nearby shop. They got together just after Valentines Day last year. She explained : 'As soon as I saw him in the shop I thought oh my god youve blossomed since school. Then I used to find reasons to go down there every day. The couple were in the same school year at Paignton Community College but only started a relationship last year. ' I went in one day and they had a collection bucket for Jack who was ill in hospital. I ran home and private messaged him on Facebook. ' Miss Cant was so concerned she decided to make the 120 mile journey to Jacks hospital in Bristol to surprise him. She said : 'He had no idea how much I liked him. 'Its not really been an ordinary relationship since day one. 'Its been hospital appointments and everything we do depends on his blood count.\\
        \hline
          \textbf{Ground truth summary:} Jack Jordan proposed to girlfriend Laura Cant while suffering leukaemia. Pair planned to marry after Mr Jordan received a bone marrow transplant. Last week he was told he was too ill to have surgery and has weeks to live. Couple have now brought forward the wedding and will marry in hospital.\\
        \midrule
        \textbf{SSTIA on clean input:} Jack Jordan proposed to his girlfriend Laura Cant last Christmas from his hospital bed. The 23-year-old had hoped to undergo a bone marrow transplant to prolong his life. But last week he was told he was too ill to undergo the operation and has just weeks to live. The couple are now set to become the first to marry in a ceremony at Torbay Hospital's chapel.
 \\
        \midrule
        \textbf{SSTIA on attacked input:} Jack Jordan, 23, proposed to his partner Laura Cant, 24, last year from his bed ward. The couple had hoped to spend their lives together after \hlc[crimsonglory!20]{a twin cancer transplant}. But Mr Jordan was given the news that he was too ill to undergo the implant and has days to live. They are now planning to tie the knot at \hlc[crimsonglory!20]{Torbase Cancer's chapel} in Bristol yesterday \\
        \midrule
        \textbf{ED on clean input:} Jack Jordan, 23, proposed to his girlfriend Laura Cant last Christmas from his hospital bed. The pair planned to wed after a bone marrow transplant which they hoped would save his life. Last week Mr Jordan was told he was too ill for the operation and has weeks to live. The couple are now set to become the first to marry in Torbay Hospital's chapel.\\
        \midrule
        \textbf{ED on attacked input:} Jack Jordan, 23, proposed to his girlfriend Laura Cant last year while battling leukaemia. The couple planned to marry after a bone marrow transplant which they hoped would give them a bright future together. But last week Mr Jordan was told he was too ill to undergo the operation and has just weeks to live. They are now planning to wed in hospital chapel at Torbay Hospital\\
     \bottomrule
        
    \end{tabular}
    \caption{The clean and attacked test input document.
    \hlc[bleudefrance!40]{Key information} in the input article and \hlc[crimsonglory!20]{inconsistent information} caused by perturbations on the baseline model is highlighted.
   }
    \label{fig:case_last}
\end{figure*}
\section{Human evaluation of Dual-view Augmentation}

We used human evaluation to assess system output on attacked test set in addition to automatic evaluation.
We employ a question-answer (QA) paradigm that assesses how well summarization algorithms maintain important information from the source text \cite{narayan2018ranking,Liu2019TextSW}.
In accordance with this paradigm, a set of questions is developed based on the gold summary on the presumption that it highlights the most crucial aspects of the document.
Then, without having access to the article, participants are required to read system summaries in order to answer to these questions.
The more questions a system can answer, the more effective it is at providing a comprehensive summary of the document.
Additionally, we evaluated the overall quality of the summaries generated by abstractive systems, considering the summarized content may produce disfluent or ungrammatical output.
Concretely, we ask two PhD students to rate according to the criteria of \textit{Informativeness}, \textit{Fluency}, and \textit{Faithfulness}.
The rating score ranges from 1 to 3, with 3 being the best. 
\cite{narayan2018ranking,Liu2019TextSW} sample 20 samples for evaluation, and we expand the size to 40 to avoid the influence of randomness.

Table~\ref{tab:comp_human_baslines} lists the average scores of each model, showing that our model outperforms other baseline models among all metrics. 
Specifically, our model outperforms BART+SSTIA by 0.38 score in terms of faithfulness and 10.35 score in QA task.
This is in accordance with the QuestEval metric, which demonstrates that our model can give more correct and salient information. 
The kappa statistics are 0.43, 0.51, and 0.49 for Info, Flu, and Faith, respectively, which indicates the moderate agreement between annotators. 
To verify the significance of these results, we also conduct the paired student t-test between our model and Sensation.
We obtain a p-value of $4 \times 10^{-4}$, $2 \times 10^{-7}$, and $6 \times 10^{-6}$ for fluency, consistency, and attractiveness.

  \begin{table}[tb]
    \centering
    \small
    \begin{tabular}{@{}lcccc@{}}
      \toprule
      & QA(\%) & Info & Flu & Faith \\
      \midrule
      BART & 40.22 &2.17  & 2.23 &2.36  \\
    BART+SSTIA & 42.52  &2.26 &2.28 & 2.38  \\
      BART+ED & \textbf{52.87} &\textbf{2.48}  & \textbf{2.43} &\textbf{2.76}  \\
      \bottomrule
    \end{tabular}
    \caption{QA accuracy, Informtiveness (Info), Fluency (Flu), and Faithfulness (Faith) comparison by human evaluation.}
    \label{tab:comp_human_baslines}
  \end{table}

\bibliography{custom}